\pdfoutput=1

\documentclass[11pt]{article}

\usepackage[preprint]{acl}

\usepackage{times}
\usepackage{latexsym}

\usepackage[T1]{fontenc}

\usepackage[utf8]{inputenc}
\usepackage{amsmath}
\usepackage{subcaption}
\usepackage{tabularx}
\usepackage{tabulary}
\usepackage{multirow}
\usepackage{lipsum, babel}
\usepackage{parskip} 
\usepackage{listings}
\usepackage[most]{tcolorbox}
\usepackage{multicol}
\usepackage{hyperref}
\usepackage{inconsolata}

\usepackage{makecell}
\usepackage{latexsym}
\usepackage{amssymb}
\usepackage{tcolorbox}
\usepackage{amsmath, amsfonts}
\usepackage{amsthm}
\usepackage{booktabs}
\usepackage[inline]{enumitem}

\usepackage{graphicx}
\usepackage{subcaption}

\usepackage{graphicx}
\usepackage{xcolor}
\usepackage{float}
\usepackage[inline]{enumitem}
\usepackage{bm}
\usepackage{soul}
\usepackage{booktabs}
\usepackage{rotating}
\usepackage{pdflscape}

\usepackage{pgfplots}
\usepackage{pgfplotstable}
\usepackage{tikz}
\usepackage{filecontents}
\pgfplotsset{compat=newest}
\usepgfplotslibrary{groupplots}

\usepackage{pdflscape} 

\usepackage{microtype}

\usepackage{inconsolata}

\usepackage{graphicx}

\title{LLMs as Data Annotators: How Close Are We to Human Performance}

\author{
Muhammad Uzair Ul Haq\textsuperscript{1, 3} \quad
Davide Rigoni\textsuperscript{2,3} \quad
Alessandro Sperduti\textsuperscript{3,4,5} \\
\textsuperscript{1}Amajor SB S.p.A, Via Noventana 192, 35027 Noventa Padovana, Italy \\
\textsuperscript{2} Department of Pharmaceutical and Pharmacological Sciences, University of Padova, Italy \\
\textsuperscript{3}Department of Mathematics ``Tullio Levi-Civita'', University of Padova, Italy \\
\textsuperscript{4}Augmented Intelligence Center, Bruno Kessler Foundation, Trento, Italy \\
\textsuperscript{5}Department of Information Engineering and Computer Science, University of Trento, Italy \\
}

\begin{document}
\maketitle

\begin{abstract}
In NLP, fine-tuning LLMs is effective for various applications but requires high-quality annotated data. However, manual annotation of data is labor-intensive, time-consuming, and costly. Therefore, LLMs are increasingly used to automate the process, often employing in-context learning (ICL) in which some examples related to the task are given in the prompt for better performance. However, manually selecting context examples can lead to inefficiencies and suboptimal model performance. 
This paper presents comprehensive experiments comparing several LLMs, considering different embedding models, across various datasets for the Named Entity Recognition (NER) task. The evaluation encompasses models with approximately $7$B and $70$B parameters, including both proprietary and non-proprietary models. Furthermore, leveraging the success of Retrieval-Augmented Generation (RAG), it also considers a method that addresses the limitations of ICL by automatically retrieving contextual examples, thereby enhancing performance. The results highlight the importance of selecting the appropriate LLM and embedding model, understanding the trade-offs between LLM sizes and desired performance, and the necessity to direct research efforts towards more challenging datasets. 

\end{abstract}


\section{Introduction}

Data annotation plays a crucial role in training machine learning (ML) models, especially in the era of Natural Language Processing (NLP). In NLP, data annotation typically involves annotating text data with relevant information, such as named entities, parts of speech, sentiment, intent, text classification, etc. The data annotation carries even more significance for fine-grained NLP tasks like token classification, where each token of a sentence has to be tagged with a gold label. 
In specialized domains such as Human Resource Management (HRM) or medical, organizations often possess large datasets that can be leveraged to enhance decision-making and operational efficiency through the use of LLM-based NLP approaches~\cite{urlana2024llmsindustriallensdeciphering}. However, for these organizations to fully harness the power of LLMs through fine-tuning, they need high-quality annotated datasets. Traditional data annotation is a labor-intensive and costly process, especially when applied to large corpora. For example, in the case of HRM, annotating a dataset of $10,000$ resumes for an information extraction task can be prohibitively time-consuming and requires significant human effort~\cite{feng-etal-2021-survey}.

Nowadays, pre-trained LLMs~\cite{BERT, RoBERTA} can be cost-effectively fine-tuned on downstream tasks. These fine-tuned models are frequently used in scenarios where continuous LLM usage for inference is too expensive, such as when using API provided by propriety services~\cite{openai2023gpt4, geminiteam2024geminifamilyhighlycapable}, or when there is the need for tailored models to meet strict performance standards while maintaining the privacy of sensitive information, such as in specialized fields~\cite{strohmeier2022handbook, Karabacak2023_medical_LLMs}. With the advent of advanced LLMs such as GPT-4~\cite{openai2023gpt4}, Qwen~\cite{qwen2.5}, and Llama~\cite{llama}, researchers and practitioners are increasingly leveraging these models to enhance the data annotation process~\cite{LLM_DA_Survey}. Pre-trained on massive corpora, LLMs offer unprecedented capabilities for automating and streamlining annotation, improving scalability, and reducing costs~\cite{wang-etal-2021-want-reduce}.

Recent studies have demonstrated that LLMs~\cite{wang2023gptnernamedentityrecognition, naraki2024augmentingnerdatasetsllms} can achieve performance comparable to human level in data annotation for Named Entity Recognition Task (NER). However, most of these evaluations are conducted on widely used benchmark datasets such as CoNLL-2003~\cite{conll2003} and WNUT-17~\cite{wnut17}. For instance, between $2023$ and $2025$, the CoNLL-2003 dataset has been utilized in $191$ studies, while WNUT-17 has been considered in $45$. In contrast, more complex datasets like SKILLSPAN~\cite{Skillspan} and GUM~\cite{GUM} have been used significantly less frequently, in only $9$ and $4$ studies\footnote{The statistics regarding the datasets usage is collected from https://paperswithcode.com/dataset/}, respectively. The results presented in this paper suggest that to gain a more comprehensive understanding of model performance regarding data annotation via LLMs in NER task, it is crucial to extend evaluations to more challenging datasets, which better reflect the complexities of real-world applications.

From a technical perspective, in the recent literature, prompting~\cite{he2024annollmmakinglargelanguage} and in-context learning (ICL)~\cite{dong2024surveyincontextlearning} are common approaches to leverage the LLMs for data annotation~\cite{LLM_DA_Survey}. ICL, which is a technique where some solved examples of the task are given within the prompt for better7 performance, is generally proven to be more effective. However, selecting the right and relevant examples to use as context for LLMs continues to be a challenging task~\cite{zhang2022automaticchainthoughtprompting}. Manually choosing examples for each query creates labor overhead, and more significantly, the use of incorrect context examples may lead LLMs to produce hallucinations~\cite{yao2024llmlieshallucinationsbugs} or inaccurate outputs.

To address the above mentioned challenges, this paper presents the following contributions:
\begin{enumerate}
    \item \textbf{Comprehensive Evaluation of LLMs and Embeddings.} It provides a comprehensive assessment of LLMs for data annotation in NER tasks, examining two distinct embedding models, as well as different techniques such as ICL and RAG, while utilizing datasets of varying complexity. It compares five models including proprietary models, such as {\tt gpt-4o-mini}, and open-source alternatives with approximately $7$B and $70$B parameters scale.
    
    \item \textbf{Trade-off Between LLM Sizes and Performances.} The trade-off between LLM sizes and performance is demonstrated, which is further verified by the statistical tests. In fact, with the appropriate LLM and embedding models, there are no statistically significant differences in results between certain $7$B and $70$B models.
    
    \item \textbf{A RAG-Based Annotation Approach.} To improve annotation quality and address the limitations of manual context selection in ICL, this paper considers a RAG based approach~\cite{RAG_NEURIPS2020_6b493230}. Instead of manually crafting in-context examples, the proposed method retrieves the most relevant samples based on similarity scores, enabling LLMs to generate more accurate annotations.

\end{enumerate}


\section{Related Work}
\label{sec:related_work}
In the recent past, there have been efforts by researchers to leverage the LLMs for data annotation~\cite{LLM_DA_Survey}. \citet{wang-etal-2021-want-reduce} introduced the use of GPT-3~\cite{brown2020languagemodelsfewshotlearners} for data annotation. The authors evaluated the quality of data generated by the GPT-3 against the human-labeled data. For each sentence to be annotated by the model, they construct a prompt consisting of several human-labeled examples along with the target sentence. They evaluate the performance in $n$-shot settings. Also, the authors report the performance of text classification and data generation tasks.
Likewise, \citet{he2024annollmmakinglargelanguage} leveraged the use of GPT-3.5 based models to annotate data. In comparison to the previous approach presented by \citet{wang-etal-2021-want-reduce}, the authors introduced the concept of chain-of-thought (CoT)~\cite{wei2023chainofthoughtpromptingelicitsreasoning} reasoning to annotate data. The authors simulate the human reasoning process to induce GPT-3.5 to motivate the annotated examples. They present the task description, specific examples, and the corresponding gold labels to GPT-3.5, and then ask the model to explain whether/why the given label is appropriate for that example. 
This enables the model to explain its choice of a specific label for the target sentence.
Then, the authors construct the few-shot CoT prompts using the explanations generated by the model for data annotation. 

To leverage the GPT model for the Named Entity Recognition (NER) task, \citet{wang2023gptnernamedentityrecognition} proposed a GPT-NER model. The main contribution introduced by the authors is to transform the NER into a text-generation task. The authors used prompt engineering, where prompts consist of three parts:
\begin{enumerate*}[label=(\it\roman*)]
    \item task description;
    \item few-shot examples; and
    \item input sentence.
\end{enumerate*}
To choose few-shot context examples, they used two different strategies:
\begin{enumerate*}[label=(\it\roman*)]
    \item random retrieval; and
    \item $k$-NN based retrieval from training data.
\end{enumerate*}

In this work, the authors propose a retrieval-based approach for selecting context examples. Specifically, for each training instance, the method iterates through all tokens in a sentence to identify the \( k \)-nearest neighbor (k-NN) tokens. The top \( k \) retrieved tokens are then selected, and their corresponding sentences are used as context. The context examples are retrieved from the entire training dataset. Furthermore, for sentences containing multiple entities, the algorithm runs multiple times to ensure the extraction of all entities within the sentence.

Following the work of \citet{wang-etal-2021-want-reduce} and \citet{wang2023gptnernamedentityrecognition}, \citet{naraki2024augmentingnerdatasetsllms} also proposed a LLMs based annotation for NER task. The authors used the LLMs to clean noise and inconsistencies in the NER dataset, and then they merged the cleaned NER dataset with the original dataset to generate a more robust and diverse set of annotations. It is worth mentioning that, in merging the annotations from LLM with human labels, preference is given to human-annotated examples compared to the LLM annotations.
In addition, \citet{bogdanov2024nunerentityrecognitionencoder} used the LLMs to create a general dataset for NER tasks with a broad range of entity types.
The authors demonstrate a procedure that consists of annotating raw data with an LLM to train a task-specific foundation model for NER.
\citet{goel2023llmsaccelerateannotationmedical} uses the same concept of data annotation using LLMs, however, they do a case study on a medical domain where they leverage the LLMs for accelerating the annotation process along with human input. 

The research discussed above highlights the strong interest in using LLMs for dataset annotation, with most approaches relying on ICL. However, systematic evaluation on complex datasets remains limited, and selecting appropriate context examples for ICL is still a challenge. This study provides a comprehensive evaluation of LLMs for NER data annotation.

\section{Methodology}
\begin{figure*}
    \centering
    {\includegraphics[width=1\linewidth]{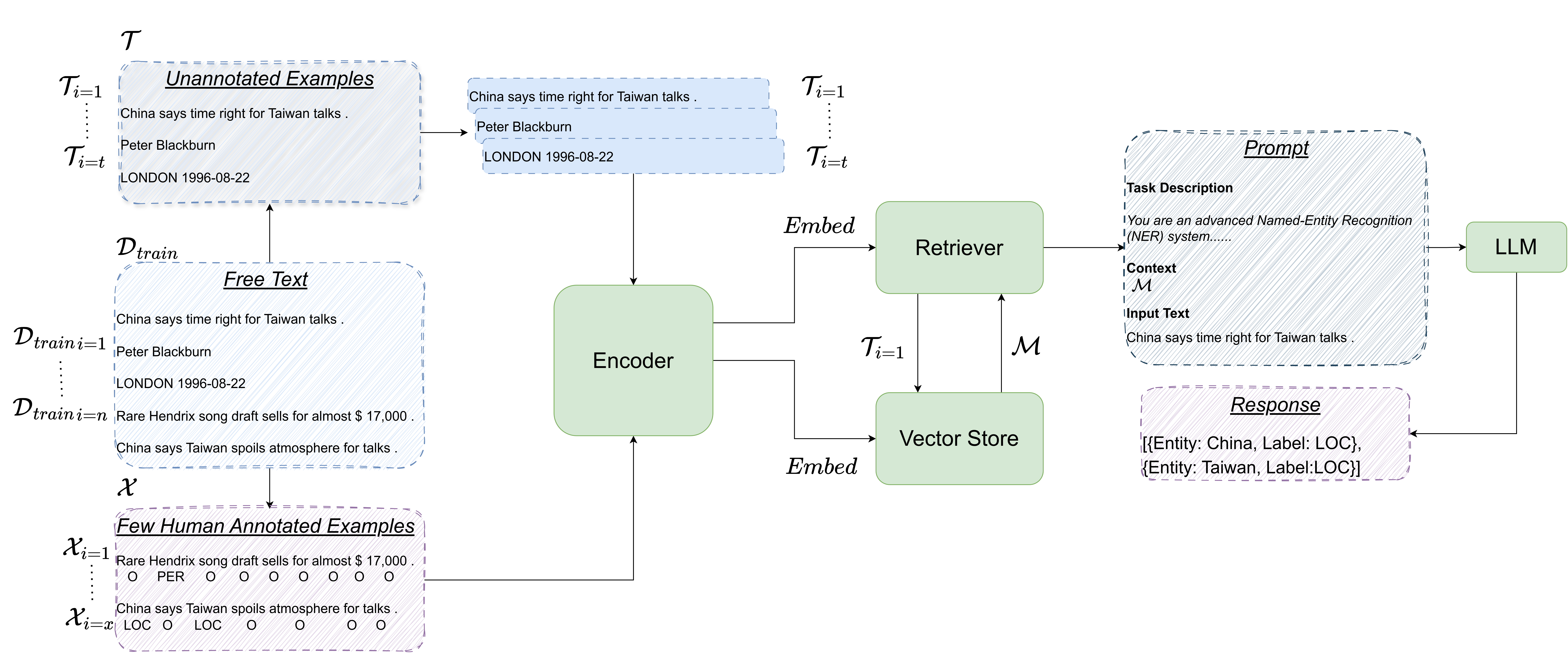}\label{fig:sub1}}
    \hspace{0.5\linewidth}
    \caption{Workflow of the proposed approach. $\mathcal{D}_{train}$ denotes the training data, $\mathcal{X}$ denotes the few human annotated examples, whereas $\mathcal{T}$ denotes the training instances to be annotated by LLM. For each entry $\mathcal{T}_i \in \mathcal{T}$, we extract $\mathcal{M}$ context examples from a vector store using a retriever module. Then, given an input sentence, the final prompt to LLM consists of the task description, the context examples in $\mathcal{M}$, and input sentence.}
    \label{fig:workflow}
\end{figure*}

\subsection{Problem Definition}
\label{problem_definition}

Given a dataset \( \mathcal{D} = \{ S_i \}_{i=1}^{n} \), where \(S_i \) represents the \( i \)-th sentence, with training, validation and test split given as \(\mathcal{D}_{train}\), \(\mathcal{D}_{valid}\) and \(\mathcal{D}_{test}\). We divide \( \mathcal{D}_{train} \) into two disjoint subsets: \( \mathcal{X} \) (we call as sample space), from which we sample context examples, and \( \mathcal{T} \), which will be annotated by the LLM. Formally, let \( \mathcal{X} \subset \mathcal{D}_{train} \) be a subset of size \( x \), where \mbox{\( x < n \)}, and \mbox{\( \mathcal{T} = \mathcal{D}_{train} \setminus \mathcal{X} \)} be the remaining subset containing \( t \) sentences, where \( t = n - x \). From \( \mathcal{X} \), we select \( m \) examples, where \( m < x \), to form the context set \( \mathcal{M} \). 
The LLM uses all the $m$ examples in \( \mathcal{M} \) as input context to annotate the $t$ sentences in \( \mathcal{T} \).

The NER task can be defined as the problem of learning an approximation function $\widetilde{f}_{\theta}$ that closely matches the real function $f: \mathcal{S}_{\mathcal{V}} \times \mathcal{V} \rightarrow \mathcal{C}$, where \( \mathcal{S}_{\mathcal{V}} \) represents the set of all the possible sentences composed only by words $w$ in the vocabulary \( \mathcal{V} \), and \( \mathcal{C} \) represents the set of possible entity categories. 
The real function $f$ given:
\begin{enumerate*}[label=(\it\roman*)]
    \item a sentence $S_i \in \mathcal{S}_{\mathcal{V}}$, and
    \item a word $w \in \mathcal{V}$,
\end{enumerate*}
assigns $w$ to its corresponding category $c \in \mathcal{C}$. 

\subsection{Data Annotation via LLMs}
\label{sec:data_annotation_strategies}
The methodology adopted in the proposed RAG approach is shown in Figure~\ref{fig:workflow}.
This section discusses the steps followed in the proposed study. Section~\ref{sec:prompt_formation} explains the prompt template formation, while Section~\ref{sec:zero_shot} presents the baseline approach, followed by ICL method in Section~\ref{sec:icl}. Section~\ref{sec:rag} presents the proposed RAG technique, whereas the importance of structured outputs for NER task is discussed in Section~\ref{sec:structured_outputs}.

\subsubsection{Prompt Formation}
\label{sec:prompt_formation}
In NLP, crafting an effective prompt for LLMs is a crucial task, as an ill-formed prompt could lead to poor performance. Different LLMs, whether open-source or proprietary, tend to respond differently to variations in prompt~\cite{errica2024didiwrongquantifying}. 
This work adopts a similar approach to prompt design presented in \cite{he-etal-2024-annollm, wang2023gptnernamedentityrecognition}, i.e. structuring our prompts around three key components, also visible in Figure~\ref{fig:workflow}:
\begin{enumerate*}[label=(\it\roman*)]
    \item \emph{Task Description}. This component clearly defines the task the LLM is expected to perform;
    \item \emph{Context}. This component provides task-related examples that help the LLM to better understand the problem, while also clarifying the expected input/output format; and
    \item \emph{Input}. This final component presents the LLM with the specific examples to be annotated.
\end{enumerate*}
The prompt structures adopted in the experiments are outlined in Appendix~\ref{app:prompt_structure}, while several prompt examples are reported in Appendix~\ref{app:examples}.

\subsubsection{Zero-shot Data Annotation}
\label{sec:zero_shot}
In the zero-shot setting (refers to the baseline), the LLM receives only task descriptions and entity categories from the dataset. The task description explains the task, whereas entity categories provide information about the classes that the LLM has to use for annotation. Providing entity categories in the prompt allows the LLM to produce consistent output annotation as in the training set. For instance, in the CoNLL-2003 dataset, $\mathsf{person}$  and $\mathsf{organization}$ categories are labelled as $\mathsf{PER}$ and $\mathsf{ORG}$ respectively. Thus, the prompt to the LLM includes $\mathsf{PER}$ and $\mathsf{ORG}$ to annotate entities in the $\mathsf{person}$ and $\mathsf{organization}$ categories, respectively. However, in zero-shot data annotation, the lack of context examples hinders the model's understanding, often leading to suboptimal performance. Nonetheless, this setting allows to evaluate the general knowledge of LLM on a task. 

\subsubsection{In-Context Learning}
\label{sec:icl}
In ICL, the prompts given to LLMs are enhanced by including not only a task description and entity categories but also contextual examples. These examples aid the models in better understanding the task at hand. As detailed in Section~\ref{problem_definition}, $\mathcal{D}_{train}$ is is split into $\mathcal{X}$ and $\mathcal{T}$. From $\mathcal{X}$, the selection of $\mathcal{M}$ can be approached in two ways: either through manual cherry-picking or by random sampling. However, manually selecting $\mathcal{M}$ can be both time-consuming and subjective, which contradicts the rationale of the proposed study. Therefore, we opt to randomly sample $\mathcal{M}$ from $\mathcal{X}$, although it does not guarantee whether the selected context examples $\mathcal{M}$ are semantically close to the input text $\mathcal{T}_{i}$, which is a limitation of this approach.

\subsubsection{Retrieval-Based Approach}
\label{sec:rag}
To overcome the limitations of the previously mentioned approaches, this paper introduces a retrieval-based method for automatically selecting relevant context examples. As outlined in Section~\ref{problem_definition}, the proposed RAG-based approach first generates embedding representations for all examples in \(\mathcal{X}\), which are then stored in a vector database~\cite{douze2024faiss} for subsequent retrieval, as illustrated in Figure~\ref{fig:workflow}.
Subsequently, for each sentence $\mathcal{T}_{i} \in \mathcal{T}$, its embedding representation is generated, and the most similar $\mathcal{M}$ examples are retrieved from $\mathcal{X}$ stored in the vector database. $\mathcal{M}$ is then used as context for the LLM to provide the most relevant examples for annotating the input text $\mathcal{T}_{i}$.

\subsubsection{Structured Output from LLMs}
\label{sec:structured_outputs}
For a label-sensitive task like NER, getting a structured output from a LLM is a crucial step. In the NER task, as defined in Section~\ref{problem_definition}, each token in a sentence is tagged with a corresponding label. 
Hence, preserving the token-label correspondence in the output is necessary for the LLMs. 
The most recent LLMs are based on a decoder architecture that, while being suitable for sequence-to-sequence tasks, encounters challenges when tackling the NER task due to the potential misalignment between tokens and labels~\cite{DA_EM}. In fact, recent studies on NER~\cite{li2024simpleeffectiveapproachimprove, LIU2024103809,wang2023gptnernamedentityrecognition} have shown that the decoder architecture presents structural inconsistencies in the output.
Recently, OpenAI~\cite{openai2023gpt4} released a feature for the latest GPT-4 based models which guarantees to follow the structured output format\footnote{\url{https://openai.com/index/introducing-structured-outputs-in-the-api}}. To solve the token-label misalignment problem, in this study, we leverage the latest feature of $\mathsf{StructuredOutput}$  released by OpenAI. 
However, it is important to note that despite the inclusion of such features in the latest LLMs, including Qwen \cite{qwen2.5} and Llama \cite{llama} based models, they still exhibit inconsistencies in their output, unlike the {\tt gpt-4o-mini-2024-07-18}. 


\section{Experimental Setup}
\subsection{Datasets}
\label{sec:dataset}

In this study, to evaluate the performance of the proposed methodology and assess the capabilities of LLMs, four datasets are considered, with their statistics summarized in Table~\ref{tab:dataset_statistics} of Appendix~\ref{app:datasets_statistics}. Each dataset presents unique challenges for LLMs in performing NER tasks, allowing this study to comprehensively analyze the ability of LLMs to handle diverse entity types, from well-structured entities to complex, ambiguous, and domain-specific annotations.

\paragraph{CoNLL-2003}

The CoNLL-2003~\cite{conll2003} dataset consists of four general entity types. Entities in this dataset typically follow structured patterns, making them relatively easier for LLMs to identify and classify.

\paragraph{WNUT-17}

The WNUT-17~\cite{wnut17} dataset contains six categories of rare entities. This dataset is particularly challenging due to its noisy text, sparse entity occurrences, and limited labeled examples per category. Improving recall on this dataset remains a significant challenge for LLMs.

\paragraph{GUM} 

The GUM~\cite{GUM} dataset is a richly annotated corpus designed for multiple NLP tasks, including NER. It captures linguistic phenomena across various domains and genres, making it a valuable resource for evaluating model performance. The dataset includes eleven distinct named entity types. Compared to CoNLL-2003 and WNUT-17, GUM presents a higher level of complexity by incorporating a diverse set of entity types spanning multiple domains.

\paragraph{SKILLSPAN} 

The SKILLSPAN~\cite{Skillspan} dataset is composed of a single entity type. Unlike traditional entities, soft skills do not follow a fixed syntactic or semantic structure, making them inherently ambiguous. These entities can range from single tokens to multi-token expressions, increasing the complexity of annotation and information extraction tasks for LLMs.

\subsection{Approaches Under Study}
In the empirical assessment of the datasets annotated by LLMs, the zero-shot data annotation approach is chosen as the baseline since it provides no context about the task to the LLM. This zero-shot setting allows the evaluation of the LLM's general knowledge of the task. 
Moreover, ICL and RAG-based approaches, detailed in Section~\ref{sec:icl} and Section~\ref{sec:rag} respectively, are considered. For both, experiments are conducted with three different numbers of context examples:
\begin{enumerate*}[label=(\it\roman*)]
    \item $25$,
    \item $50$, and
    \item $75$.
\end{enumerate*}
Experiments are conducted on a $30$\% sample of the training set \(\mathcal{D}_{\text{train}}\), while the ablation study in Appendix~\ref{app:sample_space} examines the effects of $10$\% and $20$\% sample sizes.

This paper considers five different LLMs\footnote{The models are referred to by their base names, such as \texttt{Qwen2.5-72B} for \texttt{Qwen2.5-72B-Instruct}, and so on.}:
\begin{enumerate*}[label=(\it\roman*)]
    \item {\tt gpt-4o-mini-\allowbreak{}2024-\allowbreak{}07-\allowbreak{}18},
    \item {\tt Qwen2.5-\allowbreak{}72B-\allowbreak{}Instruct},
    \item {\tt Llama3.5-\allowbreak{}70B-\allowbreak{}Instruct},
    \item {\tt Qwen2.5-\allowbreak{}7B-\allowbreak{}Instruct}, and
    \item {\tt Llama3.1-\allowbreak{}8B-\allowbreak{}Instruct},
\end{enumerate*}
and two embeddings models:
\begin{enumerate*}[label=(\it\roman*)]
    \item the {\tt text-\allowbreak{}embedding-\allowbreak{}3-\allowbreak{}large} model\footnote{https://platform.openai.com/docs/guides/embeddings}, and
    \item the {\tt sentence transformer all-MiniLM-\allowbreak{}L6-\allowbreak{}v2} model~\cite{reimers2019sentencebertsentenceembeddingsusing}.
\end{enumerate*}
Throughout the remainder of the paper, {\tt text-embedding-3-large} will be referred to as OpenAI, and {\tt sentence transformer all-MiniLM-L6-v2} will be referred to as ST.
Implementation details of results are reported Appendix~\ref{app:implementation_details}.


\subsection{NER Evaluation Process}
To assess the quality of annotations generated by LLMs, the RoBERTa model~\cite{RoBERTA} is fine-tuned on LLM-annotated datasets, leveraging its proven effectiveness in NER tasks~\cite{zhou-etal-2022-melm, Skillspan}. Initially, an LLM is employed to automatically annotate sentences in $\mathcal{T} \subset \mathcal{D}_{train}$, using strategies from Section~\ref{sec:data_annotation_strategies}. This process generates annotations for $\mathcal{T}$, resulting in a new training set, $\mathcal{\hat{T}}$, with $|\mathcal{\hat{T}}| = |\mathcal{T}|$. This annotated set is then used to fine-tune the RoBERTa model~\cite{RoBERTA}. Model selection is performed on the validation set, $\mathcal{D}_{valid}$, and the final evaluation results are based on the test set, $\mathcal{D}_{test}$. To ensure robustness and mitigate the impact of random initialization, we average the results across five different seed values. The $F_1$ score is used to assess the performances of the models.

\section{Results and Analysis}
This section presents the quantitative results of this study, as well as its analysis. Qualitative results are reported in Appendix~\ref{app:qualitative_analysis}, while Appendix~\ref{app:statistical_test} reports the statistical tests to support the findings.

\subsection{Quantitative Results}
\label{sec:quantitative_results}
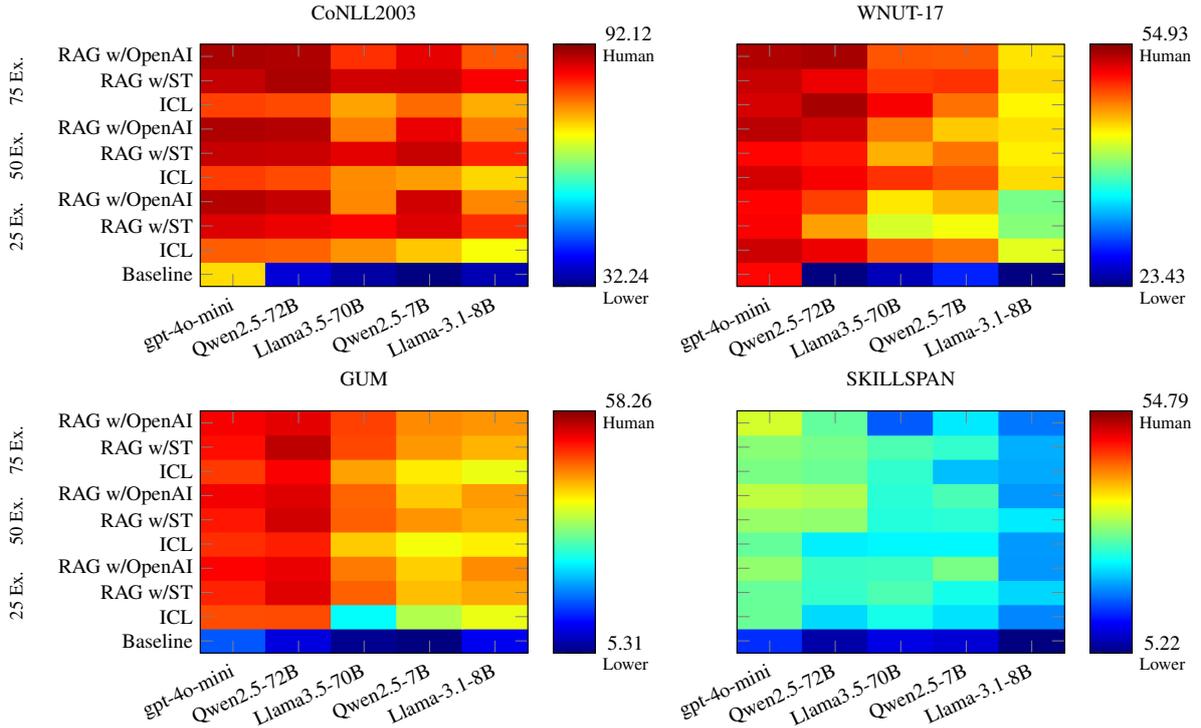
\begin{figure*}
\scriptsize
\centering
\begin{tikzpicture}[scale=1.1]
    \begin{groupplot}[
        group style={
            group size=2 by 2,
            horizontal sep=2.5cm, 
            vertical sep=1.5cm, 
            ylabels at=edge left, 
            yticklabels at=edge left,
        },
        enlarge x limits=false,
        enlarge y limits=false,
        axis on top,
        colormap/jet, 
        ytick={0,1,2,3,4,5,6,7,8,9},
        yticklabels={Baseline, ICL, RAG w/ST, RAG w/OpenAI, ICL, RAG w/ST, RAG w/OpenAI, ICL, RAG w/ST, RAG w/OpenAI},
        xtick={0,1,2,3,4},
        xticklabels={gpt-4o-mini, Qwen2.5-72B, Llama3.5-70B, Qwen2.5-7B, Llama-3.1-8B},
        xticklabel style={rotate=25, anchor=north east},
        height=4.5cm,
        width=5.5cm,
    ]
    \newcommand{\minValueFirst}{32.24}
    \newcommand{\maxValueFirst}{92.12}
    \newcommand{\minValueSecond}{23.43 }
    \newcommand{\maxValueSecond}{54.93}
    \newcommand{\minValueThird}{5.31}
    \newcommand{\maxValueThird}{58.26}
    \newcommand{\minValueFourth}{5.22}
    \newcommand{\maxValueFourth}{54.79}
    
    \nextgroupplot[
        title={CoNLL2003},
        colorbar,
        colorbar style={
            yticklabel style={
                /pgf/number format/.cd,
                fixed,
                precision=0,
                fixed zerofill,
            },
            ytick={\minValueFirst, \maxValueFirst},
            yticklabels={\shortstack{\minValueFirst\\{\tiny Lower}}, \shortstack{\maxValueFirst\\{\tiny Human}}} 
        },
        point meta min=\minValueFirst,
        point meta max=\maxValueFirst,
    ]
    \addplot [
    matrix plot*,
    point meta=explicit,
    mesh/ordering=colwise,
    ] table [x index=0, y index=1, meta index=2] {figures/heatmaps/results_dataset_1.dat};

    \nextgroupplot[
        title={WNUT-17 },
        colorbar,
        colorbar style={
            yticklabel style={
                /pgf/number format/.cd,
                fixed,
                precision=0,
                fixed zerofill,
            },
            ytick={\minValueSecond, \maxValueSecond},
            yticklabels={\shortstack{\minValueSecond\\{\tiny Lower}}, \shortstack{\maxValueSecond\\{\tiny Human}}} 
        },
        point meta min=\minValueSecond,
        point meta max=\maxValueSecond,
    ]
    \addplot [
    matrix plot*,
    point meta=explicit,
    mesh/ordering=colwise,
    ] table [x index=0, y index=1, meta index=2] {figures/heatmaps/results_dataset_2.dat};

    \nextgroupplot[
        title={GUM},
        colorbar,
        colorbar style={
            yticklabel style={
                /pgf/number format/.cd,
                fixed,
                precision=0,
                fixed zerofill,
            },
            ytick={\minValueThird, \maxValueThird},
            yticklabels={\shortstack{\minValueThird\\{\tiny Lower}}, \shortstack{\maxValueThird\\{\tiny Human}}} 
        },
        point meta min=\minValueThird,
        point meta max=\maxValueThird,
    ]
    \addplot [
    matrix plot*,
    point meta=explicit,
    mesh/ordering=colwise,
    ] table [x index=0, y index=1, meta index=2] {figures/heatmaps/results_dataset_3.dat};

    \nextgroupplot[
        title={SKILLSPAN},
        colorbar,
        colorbar style={
            yticklabel style={
                /pgf/number format/.cd,
                fixed,
                precision=0,
                fixed zerofill,
            },
            ytick={\minValueFourth, \maxValueFourth},
            yticklabels={\shortstack{\minValueFourth\\{\tiny Lower}}, \shortstack{\maxValueFourth\\{\tiny Human}}} 
        },
        point meta min=\minValueFourth,
        point meta max=\maxValueFourth,
    ]
    \addplot [
    matrix plot*,
    point meta=explicit,
    mesh/ordering=colwise,
    ] table [x index=0, y index=1, meta index=2] {figures/heatmaps/results_dataset_4.dat};
    \end{groupplot}

    \node at (-2.2cm,1.1cm) [anchor=east, rotate=90] {25 Ex.};
    \node at (-2.2cm,1.95cm) [anchor=east, rotate=90] {50 Ex.};
    \node at (-2.2cm,2.85cm) [anchor=east, rotate=90] {75 Ex.};
    \node at (-2.2cm,-1.6cm) [anchor=east, rotate=90] {75 Ex.};
    \node at (-2.2cm,-2.45cm) [anchor=east, rotate=90] {50 Ex.};
    \node at (-2.2cm,-3.35cm) [anchor=east, rotate=90] {25 Ex.};
    
\end{tikzpicture}
\caption{Heatmaps of the $F_1$ scores across four datasets. The color scale represents performance, with red indicating higher scores reaching human-level, and blue indicating lower scores starting from the lowest performing model}
\label{fig:heatmap}
\end{figure*}

Figure~\ref{fig:heatmap} presents the overall results of the experiments, while the corresponding detailed outcomes are reported in Appendix~\ref{app:detailed_results}. Specifically, the heatmaps present the $F_1$ scores obtained on the test set for different datasets, comparing several models and methods used in the proposed study. 

The CoNLL-2003 dataset, which contains named entities like persons, organizations, and locations, is relatively well-structured, making it easier for LLMs to generate high-quality annotations. The {\tt gpt-4o-mini} model with OpenAI embeddings emerges as the top performer (also shows statistical significance over other models as detailed in Appendix~\ref{app:statistical_test}), achieving an $F_1$ score of $89.72$ with $75$ context examples, which is just $2.7\%$ below human-level annotation. Among the $\sim70B$ models, {\tt Qwen2.5-72B} with OpenAI embeddings performs comparably to {\tt gpt-4o-mini} with an $F_1$ score of $89.34$, while {\tt Llama3.5-70B} with ST embeddings lags slightly behind with an $F_1$ score of $87.33$. At the $\sim 7B$ scale, {\tt Qwen2.5-7B} with ST embeddings significantly outperforms its counterparts, achieving an $F_1$ score of $87.94$, while {\tt Llama3.1-8B} with OpenAI embeddings scores $84.91$. This suggests that smaller models can still perform competitively when paired with appropriate embedding methods. Interestingly, the heatmap reveals that context size plays a crucial role—{\tt gpt-4o-mini} and {\tt Qwen2.5-70B} benefit significantly from larger context sizes of $75$ examples, while {\tt Llama3.5-70B} performs best at a slightly lower context size. This suggests that different models have varying levels of context saturation, where additional examples may not always improve performance linearly.

The WNUT-17 dataset, which focuses on low-frequency and emerging entities, presents a significant challenge due to limited training samples for each entity. However, {\tt Qwen2.5-70B} with OpenAI embeddings achieves the highest $F_1$ score of $53.72$, slightly outperforming {\tt gpt-4o-mini}, which attains an $F_1$ score of $53.43$. The {\tt Llama3.5-70B} model exhibits inconsistent performance, scoring $51.18$ with ICL at $75$ context examples, suggesting that it struggles to generalize well for rare entity detection. At the $\sim 7B$ scale, {\tt Qwen2.5-7B} with ST embeddings achieves an $F_1$ score of $49.48$, significantly outperforming {\tt Llama3.1-8B}, which scores $44.42$. This highlights that ST embeddings provide a crucial advantage for smaller models. Compared to human-level annotation, which achieves an $F_1$ score of $54.93$, the best-performing LLM reduces the gap to just $1.21\%$, which is the smallest performance gap between human and LLM annotation across all datasets used in the experiments. This suggests that RAG-based annotation is highly effective in adapting to rare entity recognition, particularly when combined with larger models and strong embeddings.

The GUM dataset presents a unique challenge due to its diverse entity types, requiring models to generalize across various linguistic structures. {\tt Qwen2.5-70B} with ST embeddings achieves the best $F_1$ score of $55.11$, significantly surpassing {\tt gpt-4o-mini}, which attains an $F_1$ score of $52.28$, and {\tt Llama3.5-70B} with OpenAI embeddings, which achieves an $F_1$ score of $48.33$. At the $\sim 7B$ scale, {\tt Qwen2.5-7B} with OpenAI embeddings achieves an $F_1$ score of $44.48$, outperforming {\tt Llama3.1-8B}, which scores $43.91$. However, both models show a notable performance drop compared to their larger counterparts, suggesting that smaller models struggle with datasets with diverse entities. The $3.15\%$ gap between the best-performing LLM and human-level annotation highlights that GUM remains a challenging dataset for LLMs. The heatmap further suggests that model performance fluctuates significantly depending on context size and embedding choice.

The SKILLSPAN dataset is the most difficult among those evaluated, as it requires understanding nuanced skill mentions across various job contexts. {\tt gpt-4o-mini} with OpenAI embeddings performs the best, achieving an $F_1$ score of $34.06$ with $75$ context examples, but this is still far from human-level annotation. At the $\sim 70B$ scale, {\tt Qwen2.5-70B} with ST embeddings achieves an $F_1$ score of $32.35$ with $50$ context examples, outperforming {\tt Llama3.5-70B}, which achieves an $F_1$ score of $27.55$. Among $\sim 7B$ models, {\tt Qwen2.5-7B} with OpenAI embeddings achieves an $F_1$ score of $29.67$, significantly surpassing {\tt Llama3.1-8B}, which scores $22.88$. This suggests that embedding choice plays a crucial role in skill extraction tasks. Notably, the gap between human annotation and the best-performing LLM is much larger in this dataset compared to others, indicating that LLMs struggle with skill-based entity recognition. This could be due to the complexity of contextual skill interpretation, requiring deeper domain knowledge and better understanding capabilities.

\subsection{Different Sample Space Choices}
\label{sec:sample_space}
\begin{figure}[h]
\centering
\begin{tikzpicture}[scale=1.0]
\label{app:figure:ablation}
    \footnotesize
    \pgfplotsset{
        x axis line style={black},
        every axis label/.append style={black},
        every tick label/.append style={black}  
    }
    \begin{axis}[
        width=0.77\linewidth,
        height=4cm,
        legend style={
            at={(1.05,0.5)}, 
            anchor=west,     
            draw=none,       
            cells={align=center}
        },
        ylabel={$\mathbf{F_1}$ score},
        ymin=26.0, ymax=33.0,
        xtick={0,1,2},
        xticklabels={\parbox{70pt}{\centering Context Size\\25},\parbox{70pt}{\centering Context Size\\50},\parbox{70pt}{\centering Context Size\\75}},
        grid=major,
        x grid style={dashed},
        y grid style={draw=none},
        point meta=explicit symbolic,
        nodes near coords,
        nodes near coords style={font=\footnotesize},
        ]
    \addplot[smooth,mark=*,color=cyan!75!black,text=black,every node near coord/.append style={yshift=+3pt}] 
    plot coordinates {
        (0,31.53)
        (1,32.07)
        (2,32.34)
    }; \addlegendentry{RAG: 10\% \hspace{0.5pt}}
    \addplot[smooth,mark=*,color=cyan!75!black,text=black,every node near coord/.append style={yshift=-12pt},dashed] 
    plot coordinates {
        (0,29.45)
        (1,30.74)
        (2,32.39)
    }; \addlegendentry{RAG: 20\% \hspace{0.5pt}}
    
    \addplot[smooth,mark=diamond*,color=orange,text=black,every node near coord/.append style={yshift=-15pt}] 
    plot coordinates {
        (0,30.74)
        (1,32.21)
        (2,31.76)
    }; \addlegendentry{ICL: 10\% \hspace{0.5pt}}
    
    \addplot[smooth,mark=diamond*,color=orange,text=black,every node near coord/.append style={yshift=+3pt},dashed] 
    plot coordinates {
        (0,27.05)
        (1,30.76)
        (2,31.32)
    }; \addlegendentry{ICL: 20\%}
    \end{axis}
    
\end{tikzpicture}
\caption{$F_1$ scores for different context sizes ($25$, $50$, and $75$) and sample spaces ($10$\% and $20$\%) for the RAG and ICL approach on the SKILLSPAN dataset, using the {\tt gpt-4o-mini} model. The plot indicates that with a smaller sample size, the RAG approach performs comparably to ICL.}
\label{fig:sample_space}
\end{figure}
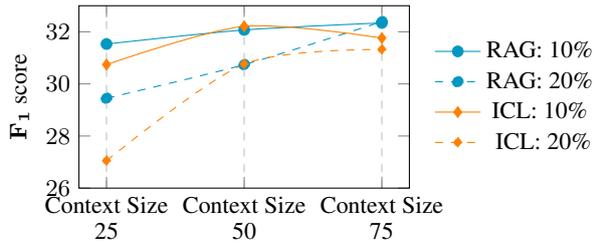

This section examines the impact of sample space choices, denoted as $\mathcal{X}$ in Section~\ref{problem_definition}, using the proposed RAG-based approach as overall it performs better than ICL. The experiments are conducted on the SKILLSPAN dataset with the {\tt gpt-4o-mini} model and OpenAI embeddings. As shown in Figure~\ref{fig:sample_space}, for smaller dataset splits, the RAG-based approach exhibits greater variability, similar to the behavior seen with ICL. This suggests that as the sample space for selecting context examples decreases, the performance of the RAG-based approach converges more closely with that of ICL. More detailed results are reported in Appendix~\ref{app:sample_space}.

\section{Discussion}
\label{sec:discussion}

\paragraph{Performance of LLMs}
The performance of different LLMs in our study reveals interesting insights. Across all datasets, RAG-based approaches improve annotation quality, with \texttt{gpt-4o-mini} and OpenAI embeddings achieving the best results. In contrast, ICL struggles in datasets with sparse or ambiguous entities, particularly SKILLSPAN. While all models perform well on CoNLL-2003, performance declines as entity structures become more complex, such as in GUM and SKILLSPAN.

\paragraph{Effect of Embeddings}
The choice of embeddings for retrieval of context for LLMs plays a crucial role in annotation quality in retrieval-based methods. OpenAI embeddings lead to better $F_1$ scores compared to smaller-scale ST embeddings especially for {\tt gpt-4o-mini} model. This effect is particularly evident in WNUT-17 and GUM, where entity distributions are more diverse, and high-quality embeddings improve retrieval effectiveness. In contrast, SKILLSPAN remains challenging across all embedding strategies, suggesting that current embedding techniques struggle with soft skill representation due to the abstract nature of the entities. 

\paragraph{Effect of Model Size}
Larger models generally perform better, but retrieval quality is equally critical. \texttt{Qwen2.5-7B} slightly outperforms \texttt{Llama3.1-8B} and performs comparably to \texttt{Llama3.5-70B} with proper embeddings, indicating that architecture and training data impact annotation beyond parameter count. Statistical tests in Appendix~\ref{app:statistical_test} support this finding.

\paragraph{Effect of Dataset Complexity}
Breaking down results per dataset, CoNLL-2003 shows minimal variance across methods, as structured entities are well-represented in training data. WNUT-17 benefits the most from retrieval-based methods, as rare entities require additional context for accurate recognition. GUM’s diverse entity types pose a challenge for ICL, but RAG-based methods significantly improve performance. Finally, SKILLSPAN remains the most difficult dataset, with lower performance across all methods, underscoring the limitations of LLMs and embeddings in capturing the semantics of soft skills.

\section{Conclusions and Future Works}

This study systematically evaluates the effectiveness of LLMs for data annotation across four diverse datasets—CoNLL-2003, WNUT-17, GUM, and SKILLSPAN of varying complexity. It compares RAG in different embedding strategies, ICL, and a baseline approach. The results demonstrate that RAG-based methods consistently outperform both ICL and the baseline across all datasets, significantly reducing the performance gap with human-level annotation.

A key finding is that dataset complexity plays a crucial role in model performance. For structured datasets like CoNLL-2003, LLMs perform exceptionally well, with models such as {\tt gpt-4o-mini} and {\tt Qwen2.5-72B} achieving results within 3\% of human-level annotation. Conversely, performance deteriorates as dataset complexity increases. The SKILLSPAN dataset, which requires nuanced skill recognition, presents the greatest challenge, with LLMs struggling to capture implicit skill mentions.

Our analysis also highlights the importance of context size and embedding choice in retrieval-augmented annotation. We observe that larger models such as  {\tt Qwen2.5-72B} and {\tt gpt-4o-mini} benefit from larger context sizes, while smaller models like {\tt Qwen2.5-7B} can still perform competitively when paired with high-quality sentence embeddings. However, models exhibit context saturation effects, where additional examples do not always lead to linear performance improvements.

Future works will focus on enhancing the performance of LLMs for complex datasets, particularly in specialized domains. In addition, future works will expand the study to more LLMs and to different NLP tasks.

\section*{Limitations}

In this study, we evaluate LLMs for data annotation tasks and introduce a RAG-based approach with different embedding models to enhance performance on NER datasets. However, our work has several limitations that highlight areas for future research.

First, our experiments focus solely on NER tasks. While this provides a solid foundation for evaluation, extending the analysis to other NLP tasks, such as text classification or question answering, would offer a more comprehensive understanding of the proposed methodology's applicability and generalizability.

Second, for the proof of concept, we employ a naïve RAG approach for context selection. Future work could explore more sophisticated retrieval techniques, such as adaptive retrieval strategies, re-ranking mechanisms, or hybrid approaches combining dense and sparse retrieval, to further optimize performance.

Third, our study does not explicitly examine the biases introduced by LLMs in the data annotation process. Given the growing concerns about fairness and model biases, a deeper investigation into how LLMs influence annotation patterns, especially in diverse and underrepresented datasets, could provide valuable insights.



\bibliography{custom, anthology.bib}


\newpage
\appendix
\onecolumn




\section{Datasets Statistics}
\label{app:datasets_statistics}
\begin{table*}[h]
\centering
\caption{Statistics of the datasets considered in this study. The average entity length refers to the average number of tokens for each entity.}
\label{tab:dataset_statistics}
\begin{tabulary}{\textwidth}{LCCCCCCC}
\toprule
\multirow{2}{*}{\textbf{Dataset}} & & \textbf{Sentences} & & & \textbf{Tokens} & & \multirow{2}{*}{\textbf{Avg. Entity Length}}  \\
\cmidrule(lr){2-4}\cmidrule(lr){5-7}
  & \textbf{Train} & \textbf{Validation} & \textbf{Test} & \textbf{Train} & \textbf{Validation} & \textbf{Test} & \\

\midrule
CoNLL-2003 & $14041$ & $3250$ & $3453$ & $203621$ & $51362$ & $46435$ & $1.60$  \\
WNUT-2017 &  $3394$   & $1008$ & $1287$ & $62730$ & $15734$ & $23394$ & $1.73$  \\
GUM &   $1435$   & $615$ & $805$ & $29392$ & $12688$ & $17437$ & $3.15$   \\
SKILLSPAN & $3074$ & $1396$ & $1522$ & $92621$ & $39923$ & $42541$  & $4.72$ \\

\bottomrule
\end{tabulary}
\end{table*}


Table~\ref{tab:dataset_statistics} highlights the complexity of entity mentions across different datasets, as reflected in their average entity length. CoNLL-2003 and WNUT-2017 contain relatively short entities, with average lengths of $1.60$ and $1.73$ tokens, respectively, indicating that most entities are single-token mentions. In contrast, GUM exhibits greater complexity, with an average entity length of $3.15$ tokens, suggesting the presence of multi-token entities. SKILLSPAN is the most complex dataset, with an average entity length of $4.72$ tokens, implying more intricate entity structures that require advanced modeling techniques for accurate recognition. 

Moreover, we discuss below the entity information for each dataset. 

\paragraph{CoNLL-2003}
The CoNLL-2003~\cite{conll2003} dataset consists of general entity types:
\begin{enumerate*}[label=(\it\roman*)]
    \item {$\mathsf{PERSON}$ }; 
    \item {$\mathsf{ORGANIZATION}$};
    \item {$\mathsf{LOCATION}$}; and 
    \item {$\mathsf{MISCELLANEOUS}$}.
\end{enumerate*}
Entities in this dataset typically follow structured patterns, making them relatively easier for LLMs to identify and classify.

\paragraph{WNUT-17}
The WNUT-17~\cite{wnut17} dataset contains six categories of rare entities:
\begin{enumerate*}[label=(\it\roman*)]
    \item {$\mathsf{PERSON}$}; 
    \item {$\mathsf{CORPORATION}$};
    \item {$\mathsf{LOCATION}$};
    \item {$\mathsf{CREATIVE\_WORK}$};
    \item {$\mathsf{GROUP}$}; and 
    \item {$\mathsf{PRODUCT}$}.
\end{enumerate*}
This dataset is particularly challenging due to its noisy text, sparse entity occurrences, and limited labeled examples per category.
\paragraph{GUM} 
The GUM~\cite{GUM} dataset is a richly annotated corpus designed for multiple NLP tasks, including NER. The dataset includes eleven distinct named entity types:
\begin{enumerate*}[label=(\it\roman*)]
    \item {$\mathsf{ABSTRACT}$}; 
    \item {$\mathsf{ANIMAL}$}; 
    \item {$\mathsf{EVENT}$}; 
    \item {$\mathsf{OBJECT}$}; 
    \item {$\mathsf{ORGANIZATION}$}; 
    \item {$\mathsf{PERSON}$}; 
    \item {$\mathsf{PLACE}$}; 
    \item {$\mathsf{PLANT}$}; 
    \item {$\mathsf{QUANTITY}$}; 
    \item {$\mathsf{SUBSTANCE}$}; and 
    \item {$\mathsf{TIME}$}.
\end{enumerate*}

\paragraph{SKILLSPAN} 
The SKILLSPAN~\cite{Skillspan} dataset is composed of a single entity type, {$\mathsf {SOFT SKILLS}$}, extracted from job descriptions. Unlike traditional entities, soft skills do not follow a fixed syntactic or semantic structure, making them inherently ambiguous.

\section{Implementation Details}
\label{app:implementation_details}

To perform experiments for data annotation with \texttt{gpt-4o-mini}, the model is accessed via the API service provided by OpenAI. To ensure reproducible results, the temperature is set to $0$ and a seed value of $42$ is used. Furthermore, the system fingerprint \texttt{fp\_1bb46167f9} is reported as noted during API access. For data annotation generation using Qwen~\cite{qwen2.5} and Llama~\cite{llama} based models, the HuggingFace \cite{HuggingFace} implementation is utilized. The instructed fine-tuned variants of the open-source models are employed in the proposed study. The models are used only for inference, with $4$-bit quantization \cite{quantization_cvpr}. The experiments with billion scale models are conducted on an $A100$ GPU with a seed value of $42$. All experiments to fine-tune NER task are performed with the RoBERTa model, available via HuggingFace~\cite{HuggingFace}, are conducted in a python environment, on an RTX $A5000$ GPU. The experiments are performed using the following five seed values: $[23112, 13215, 6465, 42, 5634]$. Moreover, the statistical significance tests are performed with the help of scikit-posthocs~\cite{statistical_test_library} library available in python.

\section{Complete Results}
\label{app:detailed_results}



\begin{landscape}
\begin{table}
\centering
\caption{The F$_1$, precision and recall along with standard deviation are reported on the test set. The values are averaged over five different random initializations. \#Ex. represents the number of context examples used. Baseline refers to the use of LLM with no context examples.}

\label{table:openai_results}
\small
\setlength{\tabcolsep}{0.4\tabcolsep}
\begin{tabular}{@{\hspace{0\tabcolsep}}l@{\hspace{6\tabcolsep}}l@{\hspace{6\tabcolsep}}ccc@{\hspace{6\tabcolsep}}ccc@{\hspace{6\tabcolsep}}ccc@{\hspace{6\tabcolsep}}ccc@{\hspace{0\tabcolsep}}}
\toprule
\multirow{2}[1]{*}{\textbf{\#Ex.}}
&\multirow{2}[1]{*}{\textbf{Method}} 
&\multicolumn{3}{c}{\textbf{CoNLL2003}} 
&\multicolumn{3}{c}{\textbf{WNUT-17}} 
&\multicolumn{3}{c}{\textbf{GUM}} 
&\multicolumn{3}{c}{\textbf{SKILLSPAN}} \\
\cmidrule(lr{0.5pt}){3-5} \cmidrule(lr{0.5pt}){6-8} \cmidrule(lr{0.5pt}){9-11} \cmidrule(lr{0.5pt}){12-14}

 &  & \textbf{P} & \textbf{R} & $\mathbf{F_1}$ & \textbf{P} & \textbf{R} & $\mathbf{F_1}$ &
\textbf{P} & \textbf{R} & $\mathbf{F_1}$ & \textbf{P} & \textbf{R} & $\mathbf{F_1}$\\
\midrule
\multirow{2}{*}{}
&Human  & $91.09_{\pm0.49}$ & $93.17_{\pm0.17}$ & $92.12_{\pm0.33}$ 
& $65.21_{\pm2.32}$ & $47.48_{\pm1.83}$ & $54.93_{\pm1.67}$  
& $55.07_{\pm0.31}$ & $61.86_{\pm0.44}$ & $58.26_{\pm0.19}$
& $54.30_{\pm1.60}$ & $55.38_{\pm1.75}$ & $54.79_{\pm0.26}$
\\

\midrule

& & & &  &\multicolumn{6}{c}{\textbf{gpt-4o-mini-2024-07-18}} & 
\\
\midrule

&Baseline & 
$64.65_{\pm0.85}$ & $80.37_{\pm0.50}$ & $71.66_{\pm0.41}$ 
& $47.35_{\pm2.46}$ & $55.18_{\pm2.84}$ & $50.88_{\pm1.14}$ 
& $20.32_{\pm5.26}$ & $13.93_{\pm2.76}$ & $16.42_{\pm3.34}$
& $11.09_{\pm0.97}$ & $17.83_{\pm2.02}$ & $13.59_{\pm0.52}$ 
\\
\midrule
\multirow{3}{*}{$25$} &ICL & 
$76.48_{\pm0.43}$ & $82.06_{\pm0.35}$ & $79.17_{\pm0.25}$ 
& $53.18_{\pm3.22}$ & $52.24_{\pm2.73}$ & $\bm{52.58}_{\pm0.78}$ 
& $44.06_{\pm0.69}$ & $52.04_{\pm1.57}$ & $47.71_{\pm0.79}$
& $21.23_{\pm1.46}$ & $45.26_{\pm1.72}$ & $28.86_{\pm1.24}$
\\
&RAG w/ST
& $84.48_{\pm1.04}$ & $88.99_{\pm0.65}$ & $86.68_{\pm0.85}$
& $51.42_{\pm2.63}$ & $50.98_{\pm1.52}$ & $51.14_{\pm1.01}$
& $46.09_{\pm0.66}$ & $54.38_{\pm1.07}$ & $49.89_{\pm0.70}$
& $20.29_{\pm0.78}$ & $49.47_{\pm1.80}$ & $28.77_{\pm0.93}$ 
\\
&RAG w/OpenAI 
& $87.35_{\pm0.65}$ & $90.71_{\pm0.34}$ & $\bm{89.00}_{\pm0.29}$ 
& $52.26_{\pm2.24}$ & $49.75_{\pm1.51}$ & $50.93_{\pm0.93}$ 
& $47.04_{\pm0.23}$ & $57.56_{\pm1.44}$ & $\bm{51.77}_{\pm0.66}$
& $21.26_{\pm1.69}$ & $56.74_{\pm1.37}$ & $\bm{30.91}_{\pm1.94}$ 
\\
\midrule
\multirow{3}{*}{$50$} &ICL 
& $79.77_{\pm0.34}$ & $82.64_{\pm0.49}$ & $81.18_{\pm0.29}$ 
& $55.75_{\pm2.80}$ & $49.53_{\pm3.07}$ & $52.33_{\pm0.97}$ 
& $45.12_{\pm0.82}$ & $54.35_{\pm2.02}$ & $49.28_{\pm0.88}$
& $20.56_{\pm0.89}$ & $47.42_{\pm2.01}$ & $28.66_{\pm0.85}$ 
\\
&RAG w/ST
& $86.73_{\pm1.03}$ & $89.29_{\pm0.84}$ & $87.99_{\pm0.90}$
& $53.74_{\pm3.02}$ & $48.74_{\pm4.44}$ & $50.90_{\pm1.34}$
& $46.46_{\pm1.34}$ & $55.46_{\pm1.21}$ & $50.56_{\pm1.29}$
& $22.22_{\pm1.47}$ & $52.60_{\pm1.41}$ & $31.20_{\pm1.32}$
\\
&RAG w/OpenAI 
& $87.43_{\pm0.48}$ & $91.39_{\pm0.16}$ & $\bm{89.36}_{\pm0.27}$ 
& $56.53_{\pm2.35}$ & $50.29_{\pm2.64}$ & $\bm{53.14}_{\pm0.75}$ 
& $47.32_{\pm0.92}$ & $58.44_{\pm1.21}$ & $\bm{52.28}_{\pm0.65}$
& $23.88_{\pm1.09}$ & $54.28_{\pm2.26}$ & $\bm{33.13_{\pm0.77}}$ 
\\
\midrule
\multirow{3}{*}{$75$} &ICL 
& $78.74_{\pm1.02}$ & $83.17_{\pm0.55}$ & $80.89_{\pm0.66}$ 
& $51.90_{\pm4.29}$ & $52.85_{\pm1.95}$ & $52.24_{\pm1.76}$ 
& $44.40_{\pm0.63}$ & $53.89_{\pm1.79}$ & $48.67_{\pm0.69}$
& $20.84_{\pm1.59}$ & $52.06_{\pm1.01}$ & $29.73_{\pm1.58}$ 
\\
&RAG w/ST
& $86.91_{\pm0.31}$ & $89.25_{\pm0.44}$ & $88.06_{\pm0.26}$
& $53.80_{\pm1.75}$ & $51.79_{\pm1.88}$ & $52.73_{\pm0.80}$
& $47.22_{\pm0.98}$ & $55.57_{\pm0.43}$ & $51.05_{\pm0.60}$
& $21.39_{\pm0.87}$ & $52.85_{\pm1.10}$ & $30.43_{\pm0.73}$
\\
&RAG w/OpenAI 
& $88.07_{\pm0.35}$ & $91.44_{\pm0.28}$ & $\bm{89.72}_{\pm0.25}$ 
& $55.72_{\pm4.22}$ & $51.71_{\pm3.34}$ & $\bm{53.43}_{\pm0.54}$ 
& $47.04_{\pm1.29}$ & $58.19_{\pm1.18}$ & $\bm{52.02}_{\pm1.15}$
& $24.66_{\pm1.34}$ & $55.39_{\pm3.19}$ & $\bm{34.06}_{\pm0.88}$ 
\\

\bottomrule

\end{tabular}
\end{table}
\end{landscape}

\begin{landscape}
\begin{table}
\centering
\caption{The F$_1$, precision and recall along with standard deviation are reported on the test set. The values are averaged over five different random initializations. \#Ex. represents the number of context examples used. Baseline refers to the use of LLM with no context examples.}

\label{table:results70B}
\small
\setlength{\tabcolsep}{0.4\tabcolsep}
\begin{tabular}{@{\hspace{0\tabcolsep}}l@{\hspace{6\tabcolsep}}l@{\hspace{6\tabcolsep}}ccc@{\hspace{6\tabcolsep}}ccc@{\hspace{6\tabcolsep}}ccc@{\hspace{6\tabcolsep}}ccc@{\hspace{0\tabcolsep}}}
\toprule
\multirow{2}[1]{*}{\textbf{\#Ex.}}
&\multirow{2}[1]{*}{\textbf{Method}} 
&\multicolumn{3}{c}{\textbf{CoNLL2003}} 
&\multicolumn{3}{c}{\textbf{WNUT-17}} 
&\multicolumn{3}{c}{\textbf{GUM}}
&\multicolumn{3}{c}{\textbf{SKILLSPAN}}  \\
\cmidrule(lr{0.5pt}){3-5} \cmidrule(lr{0.5pt}){6-8} \cmidrule(lr{0.5pt}){9-11} \cmidrule(lr{0.5pt}){12-14}

 &  & \textbf{P} & \textbf{R} & $\mathbf{F_1}$ & \textbf{P} & \textbf{R} & $\mathbf{F_1}$ &
\textbf{P} & \textbf{R} & $\mathbf{F_1}$ & \textbf{P} & \textbf{R} & $\mathbf{F_1}$\\
\midrule
\multirow{2}{*}{}
&Human  & $91.09_{\pm0.49}$ & $93.17_{\pm0.17}$ & $92.12_{\pm0.33}$ 
& $65.21_{\pm2.32}$ & $47.48_{\pm1.83}$ & $54.93_{\pm1.67}$  
& $55.07_{\pm0.31}$ & $61.86_{\pm0.44}$ & $58.26_{\pm0.19}$
& $54.30_{\pm1.60}$ & $55.38_{\pm1.75}$ & $54.79_{\pm0.26}$
\\

\midrule

&  & & & &\multicolumn{6}{c}{\textbf{Qwen2.5-72B-Instruct}} & 
\\
\midrule

&Baseline 
& $26.97_{\pm0.28}$ & $60.80_{\pm1.25}$ & $37.36_{\pm0.30}$
& $16.40_{\pm1.30}$ & $41.19_{\pm1.23}$ & $23.43_{\pm1.42}$
& $6.32_{\pm0.21}$ & $27.46_{\pm0.97}$ & $10.28_{\pm0.35}$
& $4.89_{\pm0.41}$ & $13.79_{\pm2.15}$  & $7.21_{\pm0.69}$
\\
\midrule

\multirow{3}{*}{25} 

&ICL
& $74.57_{\pm0.79}$ & $83.57_{\pm0.82}$ & $78.81_{\pm0.30}$
& $45.58_{\pm2.66}$ & $59.47_{\pm3.09}$ & $\bm{51.49}_{\pm0.92}$
& $41.69_{\pm1.10}$ & $55.80_{\pm1.00}$ & $47.73_{\pm1.06}$
& $17.06_{\pm2.18}$ & $31.17_{\pm1.63}$ & $22.01_{\pm2.13}$
\\

&RAG w/ST
& $81.87_{\pm0.72}$ & $89.90_{\pm0.46}$ & $85.69_{\pm0.56}$
& $46.55_{\pm2.70}$ & $45.68_{\pm1.43}$ & $46.06_{\pm1.33}$
& $47.79_{\pm1.05}$ & $60.15_{\pm0.99}$ & $\bm{53.26}_{\pm0.89}$
& $18.25_{\pm2.05}$ & $47.93_{\pm2.08}$ & $26.40_{\pm2.37}$
\\
&RAG w/OpenAI
 & $84.81_{\pm1.16}$ & $91.68_{\pm0.64}$ & $\bm{88.11}_{\pm0.82}$
& $48.33_{\pm2.82}$ & $49.88_{\pm1.89}$ & $49.05_{\pm1.86}$
& $47.16_{\pm0.46}$ & $59.83_{\pm0.64}$ & $52.74_{\pm0.16}$
& $18.23_{\pm1.90}$ & $50.07_{\pm4.49}$ & $\bm{26.63}_{\pm1.90}$
\\

\midrule
\multirow{3}{*}{50} 

&ICL
& $77.48_{\pm0.51}$ & $83.34_{\pm0.53}$ & $80.30_{\pm0.43}$
& $45.04_{\pm1.78}$ & $59.31_{\pm1.71}$ & $51.17_{\pm1.16}$
& $44.30_{\pm1.09}$ & $57.69_{\pm1.35}$ & $50.12_{\pm1.14}$
& $17.51_{\pm0.85}$ & $33.82_{\pm1.01}$ & $23.06_{\pm0.86}$
\\

&RAG w/ST
& $84.30_{\pm0.98}$ & $91.49_{\pm0.85}$ & $87.74_{\pm0.77}$
& $45.60_{\pm2.82}$ & $56.63_{\pm1.52}$ & $50.45_{\pm1.14}$
& $48.83_{\pm1.45}$ & $60.55_{\pm1.05}$ & $\bm{54.06}_{\pm1.25}$
& $21.32_{\pm1.82}$ & $55.79_{\pm4.56}$ & $30.84_{\pm2.52}$
\\
&RAG w/OpenAI

& $85.96_{\pm1.44}$ & $92.32_{\pm0.30}$ & $\bm{89.02}_{\pm0.66}$
& $48.66_{\pm2.91}$ & $57.09_{\pm2.42}$ & $\bm{52.46}_{\pm1.23}$
& $47.33_{\pm0.81}$ & $61.23_{\pm0.44}$ & $53.38_{\pm0.63}$
& $23.44_{\pm2.19}$ & $52.32_{\pm2.82}$ & $\bm{32.35}_{\pm2.54}$       
\\

\midrule
\multirow{3}{*}{75} 

&ICL
& $77.50_{\pm0.68}$ & $83.60_{\pm0.74}$ & $80.43_{\pm0.67}$
& $52.14_{\pm2.27}$ & $55.62_{\pm3.11}$ & $53.72_{\pm0.80}$
& $47.60_{\pm0.77}$ & $57.08_{\pm1.22}$ & $51.91_{\pm0.80}$
& $20.81_{\pm1.15}$ & $48.26_{\pm2.80}$ & $29.05_{\pm1.26}$
\\

&RAG w/ST
& $87.46_{\pm0.39}$ & $91.95_{\pm0.29}$ & $89.65_{\pm0.31}$
& $48.36_{\pm3.25}$ & $55.51_{\pm1.88}$ & $51.58_{\pm1.04}$
& $50.29_{\pm0.27}$ & $60.97_{\pm0.51}$ & $\bm{55.11}_{\pm0.17}$
& $20.99_{\pm2.12}$ & $49.99_{\pm1.23}$ & $\bm{29.52}_{\pm2.10}$
\\
&RAG w/OpenAI
& $86.77_{\pm0.54}$ & $92.05_{\pm0.72}$ & $\bm{89.34}_{\pm0.61}$
& $48.56_{\pm2.08}$ & $60.22_{\pm1.52}$ & $\bm{53.72}_{\pm0.71}$
& $47.24_{\pm1.27}$ & $60.34_{\pm0.57}$ & $52.98_{\pm0.76}$
& $19.95_{\pm0.74}$ & $50.74_{\pm1.47}$ & $28.62_{\pm0.77}$
\\

\midrule

&  & & & &\multicolumn{6}{c}{\textbf{Llama3.5-70B-Instruct}} & 
\\
\midrule

&Baseline 
& $23.56_{\pm0.10}$ & $63.25_{\pm0.17}$ & $34.33_{\pm0.15}$
& $16.35_{\pm0.74}$ & $54.65_{\pm0.42}$ & $25.16_{\pm0.84}$
& $6.44_{\pm0.08}$ & $27.79_{\pm0.35}$ & $10.46_{\pm0.13}$
& $3.51_{\pm0.08}$ & $24.30_{\pm0.64}$  & $6.14_{\pm0.12}$
\\
\midrule

\multirow{3}{*}{25} 

&ICL
& $73.59_{\pm0.78}$ & $78.73_{\pm1.03}$ & $76.06_{\pm0.41}$
& $48.77_{\pm2.20}$ & $47.66_{\pm5.18}$ & $\bm{48.00}_{\pm2.14}$
& $18.26_{\pm2.80}$ & $41.83_{\pm0.98}$ & $25.34_{\pm2.68}$
& $17.04_{\pm0.52}$ & $45.86_{\pm2.86}$ & $24.84_{\pm0.95}$\\

&RAG w/ST
& $83.15_{\pm1.42}$ & $86.37_{\pm0.90}$ & $\bm{84.72}_{\pm0.54}$
& $36.68_{\pm1.32}$ & $49.10_{\pm3.77}$ & $41.89_{\pm0.99}$
& $43.09_{\pm1.10}$ & $50.88_{\pm2.31}$ & $\bm{46.63}_{\pm0.89}$
& $\bm19.62_{\pm1.44}$ & $46.47_{\pm1.76}$ & $\bm{27.55}_{\pm1.21}$
\\
&RAG w/OpenAI
& $68.32_{\pm3.99}$ & $87.50_{\pm1.82}$ & $76.65_{\pm2.19}$
& $43.52_{\pm4.33}$ & $44.71_{\pm3.86}$ & $43.82_{\pm1.29}$
& $42.46_{\pm1.75}$ & $48.87_{\pm4.60}$ & $45.29_{\pm1.50}$
& $19.59_{\pm1.52}$ & $42.16_{\pm1.49}$ & $26.73_{\pm1.65}$
 \\
 
\midrule
\multirow{3}{*}{50} 

&ICL
 & $76.13_{\pm1.12}$ & $76.79_{\pm1.24}$ & $76.44_{\pm0.30}$
 & $50.24_{\pm2.81}$ & $48.90_{\pm2.24}$ & $\bm{49.48}_{\pm1.08}$
  & $35.67_{\pm1.83}$ & $48.79_{\pm3.19}$ & $41.12_{\pm1.07}$
 & $16.09_{\pm0.97}$ & $44.15_{\pm4.16}$ & $23.51_{\pm0.71}$
\\

&RAG w/ST
& $83.87_{\pm0.69}$ & $88.57_{\pm0.88}$ & $\bm{86.15}_{\pm0.28}$
& $42.92_{\pm2.03}$ & $48.79_{\pm2.99}$ & $45.57_{\pm0.76}$
& $43.76_{\pm1.50}$ & $50.24_{\pm2.00}$ & $\bm{46.73}_{\pm0.49}$
& $17.69_{\pm0.66}$ & $46.11_{\pm4.63}$ & $25.50_{\pm0.54}$
\\
&RAG w/OpenAI
& $68.36_{\pm1.53}$ & $89.08_{\pm0.75}$ & $77.35_{\pm0.97}$
& $44.14_{\pm1.97}$ & $51.28_{\pm2.94}$ & $47.36_{\pm0.64}$
& $43.70_{\pm2.43}$ & $49.70_{\pm1.83}$ & $46.45_{\pm1.40}$
& $18.37_{\pm2.42}$ & $44.41_{\pm4.44}$ & $\bm{25.77}_{\pm1.75}$
\\

\midrule
\multirow{3}{*}{75} 

&ICL
 & $74.94_{\pm1.03}$ & $75.15_{\pm1.03}$ & $75.04_{\pm0.70}$
 & $50.78_{\pm1.74}$ & $51.69_{\pm2.43}$ & $\bm{51.18}_{\pm1.05}$
  & $39.62_{\pm1.64}$ & $47.88_{\pm3.05}$ & $43.30_{\pm1.39}$
 & $17.55_{\pm1.05}$ & $51.80_{\pm1.68}$ & $26.19_{\pm1.14}$
\\

&RAG w/ST
& $85.70_{\pm0.60}$ & $89.03_{\pm0.55}$ & $\bm{87.33}_{\pm0.23}$
& $47.41_{\pm3.89}$ & $51.36_{\pm1.97}$ & $49.18_{\pm1.61}$
& $45.84_{\pm1.19}$ & $50.36_{\pm1.12}$ & $47.98_{\pm0.68}$
& $18.87_{\pm1.35}$ & $51.17_{\pm2.06}$ & $\bm{27.52}_{\pm1.18}$
\\
&RAG w/OpenAI 
& $76.99_{\pm1.57}$ & $87.46_{\pm1.39}$ & $81.87_{\pm0.67}$
& $49.43_{\pm4.27}$ & $48.16_{\pm5.99}$ & $48.39_{\pm2.17}$
& $44.46_{\pm0.61}$ & $52.96_{\pm1.65}$ & $\bm{48.33}_{\pm0.64}$
& $9.51_{\pm1.65}$ & $47.74_{\pm3.51}$ & $15.83_{\pm2.47}$
\\

\bottomrule

\end{tabular}
\end{table}
\end{landscape}

\begin{landscape}
\begin{table}
\centering
\caption{The F$_1$, precision and recall along with standard deviation are reported on the test set. The values are averaged over five different random initializations. \#Ex. represents the number of context examples used. Baseline refers to the use of LLM with no context examples.}
\label{table:results7B}
\small
\setlength{\tabcolsep}{0.4\tabcolsep}
\begin{tabular}{@{\hspace{0\tabcolsep}}l@{\hspace{6\tabcolsep}}l@{\hspace{6\tabcolsep}}ccc@{\hspace{6\tabcolsep}}ccc@{\hspace{6\tabcolsep}}ccc@{\hspace{6\tabcolsep}}ccc@{\hspace{0\tabcolsep}}}
\toprule
\multirow{2}[1]{*}{\textbf{\#Ex.}}
&\multirow{2}[1]{*}{\textbf{Method}} 
&\multicolumn{3}{c}{\textbf{CoNLL2003}} 
&\multicolumn{3}{c}{\textbf{WNUT-17}} 
&\multicolumn{3}{c}{\textbf{GUM}}
&\multicolumn{3}{c}{\textbf{SKILLSPAN}} \\
\cmidrule(lr{0.5pt}){3-5} \cmidrule(lr{0.5pt}){6-8} \cmidrule(lr{0.5pt}){9-11} \cmidrule(lr{0.5pt}){12-14}

 &  & \textbf{P} & \textbf{R} & $\mathbf{F_1}$ & \textbf{P} & \textbf{R} & $\mathbf{F_1}$ &
\textbf{P} & \textbf{R} & $\mathbf{F_1}$ & \textbf{P} & \textbf{R} & $\mathbf{F_1}$\\
\midrule
\multirow{2}{*}{}
&Human  & $91.09_{\pm0.49}$ & $93.17_{\pm0.17}$ & $92.12_{\pm0.33}$ 
& $65.21_{\pm2.32}$ & $47.48_{\pm1.83}$ & $54.93_{\pm1.67}$  
& $55.07_{\pm0.31}$ & $61.86_{\pm0.44}$ & $58.26_{\pm0.19}$
& $54.30_{\pm1.60}$ & $55.38_{\pm1.75}$ & $54.79_{\pm0.26}$
\\

\midrule


& & & &  &\multicolumn{6}{c}{\textbf{Qwen2.5-7B-Instruct}} & 
\\
\midrule

&Baseline 
& $21.79_{\pm1.28}$  & $62.11_{\pm0.44}$ & $32.24_{\pm1.44}$
& $20.95_{\pm2.83}$  & $44.08_{\pm3.68}$ & $28.36_{\pm3.17}$
&$3.27_{\pm0.22}$ & $14.10_{\pm1.03}$  & $5.31_{\pm0.37}$
& $5.41_{\pm1.05}$  & $35.29_{\pm2.73}$  & $9.35_{\pm1.58}$
\\
\midrule
\multirow{3}{*}{25} 

&ICL
& $70.22_{\pm1.45}$  & $75.96_{\pm1.49}$ & $72.95_{\pm0.30}$
& $47.79_{\pm3.40}$  & $47.02_{\pm2.78}$ & $\bm{47.29}_{\pm1.63}$
& $28.31_{\pm1.01}$ & $44.01_{\pm0.97}$ & $34.43_{\pm0.57}$
& $14.12_{\pm0.88}$  & $54.89_{\pm1.48}$ & $22.44_{\pm1.08}$
\\
&RAG w/ST
& $83.81_{\pm0.67}$  & $89.68_{\pm0.57}$ & $86.64_{\pm0.45}$       
& $37.82_{\pm3.45}$  & $49.68_{\pm3.12}$ & $42.81_{\pm2.14}$
& $35.90_{\pm1.86}$ & $50.09_{\pm2.04}$ & $\bm{41.80}_{\pm1.65}$
& $15.99_{\pm0.38}$  & $55.06_{\pm0.93}$ & $24.77_{\pm0.42}$
\\
&RAG w/OpenAI
& $84.05_{\pm1.15}$  & $90.85_{\pm0.31}$ & $\bm{87.32}_{\pm0.65}$
& $50.22_{\pm3.43}$  & $41.75_{\pm4.88}$ & $45.30_{\pm1.73}$
& $34.63_{\pm1.09}$ & $49.97_{\pm1.00}$ & $40.89_{\pm0.52}$
& $20.45_{\pm1.35}$  & $54.60_{\pm4.83}$ & $\bm{29.67}_{\pm1.29}$    

\\

\midrule
\multirow{3}{*}{50} 

&ICL
& $72.55_{\pm1.01}$  & $78.54_{\pm0.32}$ & $75.42_{\pm0.57}$
& $47.95_{\pm3.18}$  & $49.36_{\pm3.94}$ & $\bm{48.54}_{\pm2.51}$
& $33.51_{\pm0.76}$ & $43.59_{\pm1.18}$ & $37.88_{\pm0.56}$
& $15.13_{\pm0.90}$  & $52.64_{\pm2.98}$ & $23.47_{\pm1.01}$
\\
&RAG w/ST
& $85.78_{\pm0.69}$  & $90.21_{\pm0.38}$ & $\bm{87.94}_{\pm0.43}$
& $52.14_{\pm4.79}$  & $44.00_{\pm3.95}$ & $47.41_{\pm0.49}$
& $39.38_{\pm1.31}$ & $49.85_{\pm1.69}$ & $\bm{43.97}_{\pm0.44}$
& $17.36_{\pm1.08}$  & $51.06_{\pm2.65}$ & $25.87_{\pm1.02}$
\\
&RAG w/OpenAI
& $80.90_{\pm1.79}$  & $91.55_{\pm0.36}$ & $85.89_{\pm1.13}$
& $41.97_{\pm2.87}$  & $48.62_{\pm5.64}$ & $44.75_{\pm0.98}$
& $34.63_{\pm1.32}$ & $50.61_{\pm1.67}$ & $41.11_{\pm1.40}$
& $18.12_{\pm0.94}$  & $56.68_{\pm3.93}$ & $\bm{27.41}_{\pm0.73}$
\\

\midrule
\multirow{3}{*}{75} 

&ICL
& $81.36_{\pm1.19}$  & $75.72_{\pm1.00}$ & $78.43_{\pm0.63}$
& $47.90_{\pm5.76}$  & $47.78_{\pm3.57}$ & $47.51_{\pm1.97}$
& $34.23_{\pm2.14}$ & $46.40_{\pm1.10}$ & $39.39_{\pm1.77}$
& $12.98_{\pm1.17}$  & $51.23_{\pm6.20}$ & $20.68_{\pm1.81}$
\\
&RAG w/ST
& $86.54_{\pm1.93}$  & $88.40_{\pm1.15}$ & $\bm{87.44}_{\pm0.87}$
& $52.39_{\pm4.73}$  & $47.17_{\pm1.99}$ & $\bm{49.48}_{\pm1.30}$
& $40.14_{\pm1.39}$ & $48.15_{\pm1.09}$ & $43.76_{\pm0.73}$
& $18.34_{\pm0.46}$  & $46.32_{\pm3.08}$ & $\bm{26.25}_{\pm0.76}$
\\
&RAG w/OpenAI
& $81.67_{\pm1.51}$  & $90.96_{\pm0.30}$ & $86.06_{\pm0.88}$
& $48.73_{\pm1.31}$  & $47.85_{\pm2.49}$ & $48.25_{\pm1.30}$
& $39.56_{\pm1.25}$ & $50.87_{\pm1.25}$ & $\bm{44.48}_{\pm0.55}$
& $14.07_{\pm0.80}$  & $61.07_{\pm0.86}$ & $22.86_{\pm1.06}$
\\

\midrule

& & & & &\multicolumn{6}{c}{\textbf{Llama-3.1-8B-Instruct}} & 
\\
\midrule

&Baseline 
& $22.98_{\pm0.67}$  & $74.87_{\pm0.48}$ & $35.17_{\pm0.83}$
& $11.06_{\pm2.70}$ & $36.38_{\pm10.40}$ & $16.88_{\pm4.16}$
&$6.98_{\pm0.03}$ & $28.22_{\pm0.18}$ & $11.19_{\pm0.05}$
& $3.03_{\pm0.21}$  & $20.37_{\pm5.76}$  & $5.22_{\pm0.32}$
\\
\midrule

\multirow{3}{*}{25} 

&ICL

& $63.86_{\pm0.95}$  & $75.71_{\pm1.61}$ & $69.26_{\pm0.69}$
& $35.94_{\pm3.54}$  & $51.58_{\pm2.70}$ & $\bm{42.23}_{\pm2.38}$
& $33.95_{\pm1.97}$ & $41.74_{\pm3.02}$ & $37.39_{\pm1.85}$
& $12.40_{\pm0.85}$  & $33.63_{\pm6.65}$ & $17.93_{\pm0.66}$
\\
&RAG w/ST
& $78.44_{\pm1.18}$  & $86.16_{\pm0.88}$ & $\bm{82.11}_{\pm0.86}$
& $36.82_{\pm4.64}$  & $43.86_{\pm7.22}$ & $39.38_{\pm2.26}$
& $39.94_{\pm2.23}$ & $46.48_{\pm0.96}$ & $42.92_{\pm1.20}$
& $14.95_{\pm1.88}$  & $42.25_{\pm6.45}$ & $\bm{21.87}_{\pm1.51}$
\\
&RAG w/OpenAI
& $69.03_{\pm1.02}$  & $86.41_{\pm2.27}$ & $76.73_{\pm1.02}$
& $32.83_{\pm3.20}$  & $48.82_{\pm7.41}$ & $38.89_{\pm2.78}$
& $40.77_{\pm1.82}$ & $49.07_{\pm2.80}$ & $\bm{44.45}_{\pm0.54}$
& $12.16_{\pm0.97}$  & $41.55_{\pm3.20}$ & $18.79_{\pm1.31}$
\\

\midrule
\multirow{3}{*}{50} 

&ICL
& $67.78_{\pm1.48}$  & $76.79_{\pm0.69}$ & $72.01_{\pm1.13}$
& $40.49_{\pm1.76}$  & $48.82_{\pm2.88}$ & $\bm{44.20}_{\pm1.03}$
& $36.43_{\pm1.51}$ & $42.53_{\pm2.12}$ & $39.22_{\pm1.32}$
& $12.94_{\pm1.13}$  & $35.45_{\pm2.12}$ & $18.90_{\pm1.01}$
\\
&RAG w/ST
& $79.29_{\pm3.86}$  & $86.85_{\pm2.00}$ & $\bm{82.82}_{\pm1.41}$
& $40.04_{\pm4.26}$  & $48.60_{\pm4.26}$ & $43.59_{\pm1.04}$
& $39.73_{\pm1.81}$ & $46.54_{\pm2.12}$ & $42.81_{\pm0.99}$
& $15.13_{\pm0.96}$  & $47.13_{\pm1.48}$ & $\bm{22.88}_{\pm0.99}$
\\
&RAG w/OpenAI
& $69.98_{\pm1.50}$  & $87.05_{\pm2.15}$ & $77.56_{\pm0.74}$
& $39.75_{\pm1.82}$  & $49.53_{\pm2.63}$ & $44.03_{\pm0.76}$
& $40.89_{\pm1.62}$ & $46.96_{\pm2.27}$ & $\bm{43.66}_{\pm0.67}$
& $12.09_{\pm1.18}$  & $42.63_{\pm3.13}$ & $18.76_{\pm1.19}$
\\

\midrule
\multirow{3}{*}{75} 

&ICL
& $71.44_{\pm2.02}$  & $77.86_{\pm2.41}$ & $74.47_{\pm1.00}$
& $39.13_{\pm1.16}$  & $48.70_{\pm2.37}$ & $43.35_{\pm0.84}$
& $34.33_{\pm1.27}$ & $41.30_{\pm2.37}$ & $37.43_{\pm0.53}$
& $13.37_{\pm1.12}$  & $37.83_{\pm4.68}$ & $19.67_{\pm1.20}$
\\
&RAG w/ST
& $82.36_{\pm2.15}$  & $87.69_{\pm1.70}$ & $\bm{84.91}_{\pm0.88}$
& $39.92_{\pm2.09}$  & $50.34_{\pm3.12}$ & $\bm{44.42}_{\pm0.59}$
& $41.14_{\pm1.47}$ & $43.71_{\pm3.36}$ & $42.30_{\pm1.57}$
& $12.77_{\pm0.10}$  & $45.34_{\pm0.80}$ & $\bm{19.92}_{\pm0.08}$
\\
&RAG w/OpenAI 
& $74.58_{\pm2.51}$  & $85.21_{\pm0.99}$ & $79.51_{\pm1.02}$
& $41.85_{\pm2.52}$  & $47.17_{\pm6.92}$ & $43.96_{\pm2.54}$
& $41.99_{\pm1.01}$ & $46.13_{\pm2.68}$ & $\bm{43.91}_{\pm0.93}$
& $10.42_{\pm0.96}$  & $49.98_{\pm3.64}$ & $17.22_{\pm1.29}$
\\

\bottomrule

\end{tabular}
\end{table}
\end{landscape}

\section{Further Results on Different Sample Space Choices}
\label{app:sample_space}

Tables~\ref{table:sample_space} examine the influence of sample space \(\mathcal{X}\) and context size \(\mathcal{M}\) on entity recognition performance using the best-performing model, \texttt{gpt-4o-mini}, on the SKILLSPAN dataset. Increasing the context size from $25$ to $75$ generally improves the $F_1$ score, though gains diminish beyond $50$ examples. RAG consistently outperforms ICL in recall and $F_1$ score, demonstrating its effectiveness in leveraging external knowledge, while ICL achieves higher precision but lower recall, suggesting a more conservative prediction approach. At a $10$\% sample space, ICL delivers competitive results, but as it increases to $20$\%, RAG maintains a clear advantage, achieving the highest $F_1$ score of $32.39$\% at a context size of $75$. Notably, for smaller dataset splits, RAG exhibits greater variability, similar to ICL, suggesting that when fewer examples are available, their performances converge. These findings underscore the importance of context size and external knowledge availability in optimizing RAG-based methods. 


\begin{table}[h]
\centering
\small
\caption{Study comparing RAG and ICL methods at different size of sample spaces (10\% and 20\%) and context sizes (25, 50, and 75). Experiments were conducted on the SKILLSPAN dataset using the gpt-4o-mini-2024-07-18 model. The results are presented with standard deviations, showing how performance metrics vary across sampling choices and context sizes for both methods.}
\label{table:sample_space}
\begin{tabular}{ccccc
}
\toprule
\textbf{Sample Space} & \textbf{Context Size} & \textbf{Precision} & \textbf{Recall} & \textbf{F1 Score} \\
\midrule
\multicolumn{5}{c}{\textbf{RAG}} \\
\midrule
\multirow{3}{*}{10\%} 
& $25$ & $21.83_{\pm1.22}$ & $56.94_{\pm1.17}$ & $31.53_{\pm1.18}$ \\ 
& $50$ & $22.44_{\pm1.32}$ & $56.46_{\pm2.46}$ & $32.07_{\pm1.13}$ \\
& $75$ & $22.82_{\pm0.58}$ & $55.82_{\pm1.40}$ & $\bm{32.34}_{\pm0.41}$ \\
\midrule
\multirow{3}{*}{20\%} 
& $25$ & $20.26_{\pm1.55}$ & $54.46_{\pm3.71}$ & $29.45_{\pm1.29}$ \\ 
& $50$ & $21.00_{\pm0.81}$ & $57.40_{\pm1.57}$ & $30.74_{\pm0.86}$ \\
& $75$ & $22.69_{\pm0.46}$ & $56.31_{\pm2.06}$ & $\bm{32.39}_{\pm0.60}$ \\
\midrule
\multicolumn{5}{c}{\textbf{ICL}} \\
\midrule
\multirow{3}{*}{10\%} 
& $25$ & $22.57_{\pm1.49}$ & $48.72_{\pm3.95}$ & $30.74_{\pm0.87}$ \\ 
& $50$ & $23.62_{\pm0.85}$ & $50.73_{\pm1.33}$ & $\bm{32.21}_{\pm0.67}$ \\
& $75$ & $23.12_{\pm1.16}$ & $51.09_{\pm4.19}$ & $31.76_{\pm0.89}$ \\
\midrule
\multirow{3}{*}{20\%} 
& $25$ & $19.35_{\pm1.57}$ & $45.83_{\pm4.74}$ & $27.05_{\pm0.66}$ \\ 
& $50$ & $22.02_{\pm1.56}$ & $51.17_{\pm1.15}$ & $30.76_{\pm1.39}$ \\
& $75$ & $22.89_{\pm1.11}$ & $49.78_{\pm2.89}$ & $\bm{31.32}_{\pm0.99}$ \\
\bottomrule
\end{tabular}
\end{table}

\section{Statistical Significance Test}
\label{app:statistical_test}
This study evaluated various large language models across multiple datasets, considering different embeddings and examples as context. While some models clearly outperformed others in the results, the differences in predictions might not be statistically significant for certain models. Therefore, to determine the statistical significance of our findings, we conducted a non-parametric test. This test helps us assess whether there are significant differences among the models and, if so, identify which models differ statistically from each other.

The Friedman test~\cite{friedman_test} is a non-parametric statistical test used to detect differences in performance across multiple related samples --- in this case, different models evaluated over multiple datasets. It ranks the performance scores among datasets and assesses whether the rank distributions differ significantly among models. Let $N$ be the number of datasets, $K$ the number of models, and $R_j$ be the sum of ranks for each model $j$. The Friedman test statistic $chi^2_F$, which follows a chi-square distribution, is calculated as follows:
\begin{equation}
    \chi^2_F = \frac{12N}{k(k+1)} \sum_{j=1}^{k} R_j^2 - 3N(k+1).
\end{equation}
If the test statistic exceeds the critical value for a significance level $\alpha=0.01$, we reject the null hypothesis, indicating that there are significant differences in performance among the models. If significant differences are found, the post-hoc Conover~\cite{conover1999practical_post_hoc} test is performed to discover pair-wise statistical differences among models while adjusting for multiple comparisons. This test evaluates whether specific models differ significantly in performance.

\begin{figure}
    \centering
    {\includegraphics[width=1\linewidth]{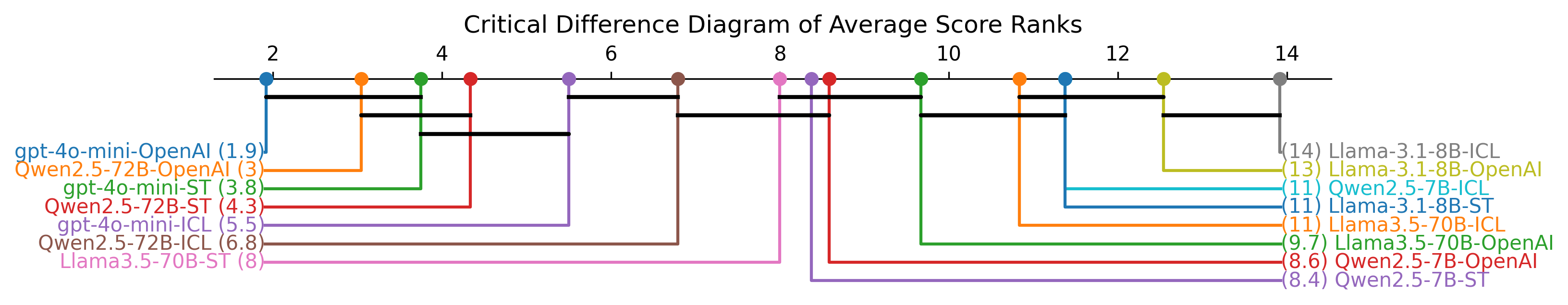}}
    \caption{Critical Difference diagram of average score ranks. The models connected with horizontal line shows no statistical difference. The models with lower ranks shows superior performance than those of higher ranks.}
    \label{fig:statistica_test}
\end{figure}

Given that the Friedman test produces a test statistic of $114.42$ with a $p$-value of $7.71^{-18}$, we reject the null hypothesis, suggesting that at least one model shows a statistically significant difference in performance. Consequently, we conducted the post-hoc Conover test. 
Figure~\ref{fig:statistica_test} presents the statistical significance of the model rankings, with significant pairwise differences highlighted accordingly.
The x-axis indicates the average rank of each model, where lower ranks closer to the left signify better performance. Each colored node corresponds to a particular model, labeled with its respective rank, while the black horizontal bars connecting multiple nodes highlight groups of models that do not show statistically significant differences at the specified confidence level.
The top-performing combination is gpt4omini-OpenAI, with an average rank of $1.9$, indicating it consistently outperformed other approaches. Other strong performers include {\tt Qwen2.5-72B-OpenAI} ($3$), {\tt gpt-4o-mini-ST} ($3.8$), and {\tt Qwen2.5-72B-ST} ($4.3$). These models have lower rankings and are clustered towards the left. In contrast, {\tt Llama3.1-8B-ICL} ($14$), {\tt Llama3.1-8B-OpenAI} ($13$), and {\tt Qwen2.5-7B-ICL} ($11$) have the highest ranks, suggesting they performed the worst in comparison. These models do not overlap with the higher-ranked ones, highlighting their statistically inferior performance.
Interestingly, {\tt Llama3.1-8B-ST} shows no statistical differences when compared to {\tt Llama3.5-70B}, whether using ICL or RAG with OpenAI embedding. Similarly, 
{\tt Qwen2.5-7B}, when utilizing RAG with either OpenAI or ST embeddings, exhibits no statistical differences compared to {\tt Llama3.5-70B} using ST embeddings and {\tt Qwen2.5-72B} using ICL. These tests highlight a crucial aspect: a trade-off when addressing the NER task. Indeed, larger models, such as those with $70$B parameters, may not necessarily offer better performance than smaller models like {Llama3.1-8B-ST} or {\tt Qwen2.5-7B}. This suggests that the additional computational resources required for bigger models might not always justify their use, especially if smaller models can achieve statistically similar results.

\section{Qualitative Analysis}
\label{app:qualitative_analysis}
\setlength{\extrarowheight}{4pt}
\begin{table*}[t]
\small
\caption{Qualitative analysis of soft skills annotations on dataset samples using \texttt{gpt-4o-mini-2024-07-18}. The output of the best-performing model is reported.
The highlighted texts in the first column are gold labels, while those in the other columns are the corresponding LLM-generated annotations.}

\label{tab:qualitative_analysis}

\begin{tabulary}{\textwidth}{CJJJJ} 
\toprule

\textbf{\textnumero} &
  \multicolumn{1}{c}{ \textbf{Human}} &
  \multicolumn{1}{c}{
  \textbf{Baseline}} &
  \multicolumn{1}{c}{
  \textbf{ICL}} &
  \multicolumn{1}{c}{
  \textbf{RAG}}  \\ 
  
\midrule
  
1. & Very good understanding of test automation frameworks.
&
\hl{Very} \hl{good} \hl{understanding} of \hl{test} \hl{automation} \hl{frameworks}.
&
Very good \hl{understanding} \hl{of} \hl{test} \hl{automation} \hl{frameworks}.
&
Very good \hl{understanding} \hl{of} \hl{test} \hl{automation} \hl{frameworks}.
\\

2. &
Must have excellent \hl{verbal} \hl{and} \hl{written} \hl{skills} being able to \hl{communicate} \hl{effectively} \hl{on} \hl{both} \hl{a} \hl{technical} \hl{and} \hl{business} \hl{level} Ability to \hl{work} \hl{under} \hl{pressure} to \hl{resolve} \hl{issues} \hl{affecting} \hl{the} \hl{production} \hl{services}.
&
Must have \hl{excellent} \hl{verbal} \hl{and} \hl{written} \hl{skills} being able to \hl{communicate} \hl{effectively} on both a technical and business level \hl{Ability} to \hl{work} \hl{under} \hl{pressure} to \hl{resolve} \hl{issues} affecting the production services. 
&
Must have \hl{excellent} \hl{verbal} \hl{and} \hl{written} \hl{skills} being able to \hl{communicate} \hl{effectively} \hl{on} \hl{both} \hl{a} \hl{technical} \hl{and} \hl{business} \hl{level} \hl{Ability} \hl{to} \hl{work} \hl{under} \hl{pressure} \hl{to} \hl{resolve} \hl{issues} affecting the production services. 
&
Must have excellent \hl{verbal} \hl{and} \hl{written} \hl{skills} being able to \hl{communicate} \hl{effectively} \hl{on} \hl{both} \hl{a} \hl{technical} \hl{and} \hl{business} \hl{level} Ability to \hl{work} \hl{under} \hl{pressure} to \hl{resolve} \hl{issues} \hl{affecting} \hl{the} \hl{production} \hl{services}.
 \\
 
3. &
Must have excellent \hl{work} \hl{ethic} and be \hl{detail} \hl{oriented} and be able to \hl{work} \hl{independently}. 
&
Must have \hl{excellent} \hl{work} \hl{ethic} and be \hl{detail} \hl{oriented} and be able to \hl{work} \hl{independently}. 
&
Must have \hl{excellent} \hl{work} \hl{ethic} and be \hl{detail} \hl{oriented} and be able to \hl{work} \hl{independently}.
&
Must have \hl{excellent} \hl{work} \hl{ethic} and be \hl{detail} \hl{oriented} and be able to \hl{work} \hl{independently}.
\\

4. &
Technical Skills Core Java. 
&
\hl{Technical} \hl{Skills} \hl{Core} \hl{Java}. 
&
Technical Skills \hl{Core} \hl{Java}.
&
Technical Skills \hl{Core} \hl{Java}.
 \\
 
5. & You will \hl{work} \hl{with} \hl{the} \hl{business} to \hl{define} \hl{requirements} and have excellent \hl{communication} \hl{skills} to interpret these into consolidated development scopes.
&
You will work with the business to define requirements and have \hl{excellent} \hl{communication} \hl{skills} to interpret these into consolidated development scopes. 
&
You will work with the business to define requirements and have \hl{excellent} \hl{communication} \hl{skills} to interpret these into consolidated development scopes. 
&
You will work with the business to \hl{define} \hl{requirements} and have \hl{excellent} \hl{communication} \hl{skills} to \hl{interpret} \hl{these} \hl{into} \hl{consolidated} \hl{development} \hl{scopes}.
\\
 \bottomrule

\end{tabulary}
\end{table*}

This study broadly explores the efficacy of LLMs for data annotation tasks. Four different datasets of varying complexity are chosen. From Table~\ref{fig:heatmap}, it is observed that the performance of LLMs decreases as dataset complexity increases. The performance of LLMs on the SKILLSPAN dataset is significantly lower than human annotation, suggesting that even the latest available LLMs struggle to annotate data when the task is complex. For instance, soft skills lack clear or distinct definitions, making the task more challenging. Similarly, the GUM dataset also poses challenges for LLMs due to its entity diversity. On the other hand, in the case of the WNUT-17 and CoNLL-2003 datasets, which consist of simpler entities (more details are reported in Section~\ref{sec:dataset}), annotations are easier to extract for an LLM given its prior knowledge. Furthermore, the quality of context in LLMs plays a major role, particularly in data annotation tasks, as indicated by Tables~\ref{table:openai_results},~\ref{table:results70B}, and~\ref{table:results7B}, where the RAG-based approach significantly outperforms its counterpart. Moreover, for simpler datasets, the RAG-based approach achieves performance comparable to human annotation.

To gain better insights into the performance of the proposed RAG-based approach, Table~\ref{tab:qualitative_analysis} presents the qualitative results for the SKILLSPAN dataset annotated by {\tt gpt-4o-mini}. In this dataset, data annotation performance remains far below human-level, suggesting that the LLM struggles to extract sufficient information from the context examples when the task is difficult. From Tables~\ref{table:openai_results},~\ref{table:results70B}, and~\ref{table:results7B}, it is observed that LLM-generated annotations improve recall, whereas precision is compromised. Table~\ref{tab:qualitative_analysis} shows that in examples $1$ and $4$, the LLM incorrectly annotates soft skills that are not identified by human annotators, whereas in examples $2$ and $3$, the annotations are nearly identical to human annotations. In Example $5$, the RAG-based approach performs comparably to human annotation, while both the baseline and ICL fail to do so.

\section{Prompt}
\label{app:prompt_structure}

This section presents the prompts used to generate the response of LLMs. These prompts are carefully synthesized to encompass all the components required to get structured output for both:
\begin{enumerate*}[label=(\it\roman*)]
    \item baseline, and
    \item in-context learning models.
\end{enumerate*}

\begin{tcolorbox}[colback=gray!5!white,colframe=gray!75!black,title=Baseline Prompt Structure]
\textbf{Task Description}
\small

You are an advanced Named-Entity Recognition (NER) system.

Your task is to analyze the given sentence or passage, identify, extract, and classify specific named entities according to the following predefined entity types:
\begin{itemize}
    \item \{labels\} 
\end{itemize}

For each sentence:
\begin{itemize}
    \item \textbf{Label} each word in the text with the appropriate entity type if it matches the specified categories.
    \item Extract \textbf{multiple entities} of the same class if they exist.
\end{itemize}

The output should be in \textbf{valid JSON format}, with each word and its corresponding label as shown below.

Follow the structure strictly and do not add any other explanation.

In entities, label the word exactly as in the text. All the text is case-sensitive.

\subsection*{Input}

\{input\_text\}
\end{tcolorbox}


\begin{tcolorbox}[colback=gray!5!white,colframe=gray!75!black,title=Context Prompt Structure]
\small

\textbf{Task Description}

You are an advanced Named-Entity Recognition (NER) system.

Your task is to analyze the given sentence or passage, identify, extract, and classify specific named entities according to the following predefined entity types:

\begin{itemize}
    \item \{labels\}
\end{itemize}

For each sentence:
\begin{itemize}
    \item \textbf{Label} each word in the text with the appropriate entity type if it matches the specified categories.
    \item Extract \textbf{multiple entities} of the same class if they exist.
\end{itemize}

The output should be in \textbf{valid JSON format}, with each word and its corresponding label as shown below.

Follow the structure strictly and do not add any other explanation.

In entities, label the word exactly as in the text. All the text is case-sensitive.

\subsection*{Examples}

\{context\_examples\}

\subsection*{Input}

\{input\_text\}
\end{tcolorbox}

\section{Examples}
\label{app:examples}

This section provides examples of prompts from the training data for different datasets used in this study. For visual purposes, we used only only $top 5$ examples in context.
Follows several prompt examples for the:
\begin{enumerate*}[label=(\it\roman*)]
    \item CoNLL-2003, 
    \item WNUT-17,  
    \item SKILLSPAN datasets, and 
    \item GUM datasets.
\end{enumerate*}

\small

\begin{tcolorbox}[colback=gray!5!white, colframe=gray!75!black, title=Example 1--CoNLL-2003, sharp corners=southwest, breakable]
\label{appendixB-CoNLL-2003-1}
\small
\textbf{Task Description}

You are an advanced Named-Entity Recognition (NER) system. Your task is to analyze the given sentence or passage, identify, extract, and classify specific named entities according to the following predefined entity types:
\\

\textbf{['PER', 'ORG', 'LOC', 'MISC']}
\\

For each sentence:
\begin{itemize}
    \item \textbf{Label} each word in the text with the appropriate entity type if it matches the specified categories.
    \item Extract \textbf{multiple entities} of the same class if they exist.
\end{itemize}

The output should be in \textbf{valid JSON format}, with each word and its corresponding label as shown below.

Follow the structure strictly and do not add any other explanation. In entities, label the word exactly as in the text. All the text is case-sensitive.

\subsection*{Examples}

\begin{lstlisting}[breaklines]
    
["A South African boy is writing back to an American girl whose message in a bottle he found washed up on President Nelson Mandela 's old prison island .", [{'Entity': 'South African', 'Label': 'MISC'}, {'Entity': 'American', 'Label': 'MISC'}, {'Entity': 'Nelson Mandela', 'Label': 'PER'}]]

['A rottweiler dog belonging to an elderly South African couple savaged to death their two-year-old grandson who was visiting , police said on Thursday .', [{'Entity': 'South African', 'Label': 'MISC'}]]

['The princess , who has carved out a major role for herself as a helper of the sick and needy , is said to have turned to Mother Teresa for guidance as her marriage crumbled to heir to the British throne Prince Charles .', [{'Entity': 'Mother Teresa', 'Label': 'PER'}, {'Entity': 'British', 'Label': 'MISC'}, {'Entity': 'Prince Charles', 'Label': 'PER'}]]

['South African answers U.S. message in a bottle .', [{'Entity': 'South African', 'Label': 'MISC'}, {'Entity': 'U.S.', 'Label': 'LOC'}]]

["But Carlo Hoffmann , an 11-year-old jailer 's son who found the bottle on the beach at Robben Island off Cape Town after winter storms , will send his letter back by ordinary mail on Thursday , the post office said .", [{'Entity': 'Carlo Hoffmann', 'Label': 'PER'}, {'Entity': 'Robben Island', 'Label': 'LOC'}, {'Entity': 'Cape Town', 'Label': 'LOC'}]]

\end{lstlisting}

\subsection*{Input}

Revered skull of S. Africa king is Scottish woman 's .
\\

\hrule

\subsection*{Response}

[{Entity: S. Africa, Label: LOC}, {Entity: Scottish, Label: MISC}]

\end{tcolorbox}



\begin{tcolorbox}[colback=gray!5!white, colframe=gray!75!black, title=Example 2--CoNLL-2003, sharp corners=southwest, breakable]
\label{appendixB-CoNLL-2003-2}
\small
\textbf{Task Description}

You are an advanced Named-Entity Recognition (NER) system. Your task is to analyze the given sentence or passage, identify, extract, and classify specific named entities according to the following predefined entity types:
\\

\textbf{['PER', 'ORG', 'LOC', 'MISC']}
\\

For each sentence:
\begin{itemize}
    \item \textbf{Label} each word in the text with the appropriate entity type if it matches the specified categories.
    \item Extract \textbf{multiple entities} of the same class if they exist.
\end{itemize}

The output should be in \textbf{valid JSON format}, with each word and its corresponding label as shown below.

Follow the structure strictly and do not add any other explanation. In entities, label the word exactly as in the text. All the text is case-sensitive.

\subsection*{Examples}

\begin{lstlisting}[breaklines]
    
['Rwanda said on Saturday that Zaire had expelled 28 Rwandan Hutu refugees accused of being " trouble-makers " in camps in eastern Zaire .', [{'Entity': 'Rwanda', 'Label': 'LOC'}, {'Entity': 'Zaire', 'Label': 'LOC'}, {'Entity': 'Rwandan', 'Label': 'MISC'}, {'Entity': 'Hutu', 'Label': 'MISC'}, {'Entity': 'Zaire', 'Label': 'LOC'}]]

['Repatriation of 1.1 million Rwandan Hutu refugees announced by Zaire and Rwanda on Thursday could start within the next few days , an exiled Rwandan Hutu lobby group said on Friday .', [{'Entity': 'Rwandan Hutu', 'Label': 'MISC'}, {'Entity': 'Zaire', 'Label': 'LOC'}, {'Entity': 'Rwanda', 'Label': 'LOC'}, {'Entity': 'Rwandan Hutu', 'Label': 'MISC'}]]

['Innocent Butare , executive secretary of the Rally for the Return of Refugees and Democracy in Rwanda ( RDR ) which says it has the support of Rwanda \'s exiled Hutus , appealed to the international community to deter the two countries from going ahead with what it termed a " forced and inhuman action " .', [{'Entity': 'Innocent Butare', 'Label': 'PER'}, {'Entity': 'Rally for the Return of Refugees and Democracy in Rwanda', 'Label': 'ORG'}, {'Entity': 'RDR', 'Label': 'ORG'}, {'Entity': 'Rwanda', 'Label': 'LOC'}, {'Entity': 'Hutus', 'Label': 'MISC'}]]

['Rwanda says Zaire expels 28 Rwandan refugees .', [{'Entity': 'Rwanda', 'Label': 'LOC'}, {'Entity': 'Zaire', 'Label': 'LOC'}, {'Entity': 'Rwandan', 'Label': 'MISC'}]]

['Rwandan group says expulsion could be imminent .', [{'Entity': 'Rwandan', 'Label': 'MISC'}]]

\end{lstlisting}

\subsection*{Input}

Captain Firmin Gatera , spokesman for the Tutsi-dominated Rwandan army , told Reuters in Kigali that 17 of the 28 refugees handed over on Friday from the Zairean town of Goma had been soldiers in the former Hutu army which fled to Zaire in 1994 after being defeated by Tutsi forces in Rwanda 's civil war .
\\

\hrule

\subsection*{Response}

[{Entity: Captain Firmin Gatera, Label: PER}, {Entity: Rwandan, Label: MISC}, {Entity: Reuters, Label: ORG}, {Entity: Kigali, Label: LOC}, {Entity: Zairean, Label: MISC}, {Entity: Goma, Label: LOC}, {Entity: Hutu, Label: MISC}, {Entity: Zaire, Label: LOC}, {Entity: Tutsi, Label: MISC}, {Entity: Rwanda, Label: LOC}]
 
 \end{tcolorbox}


\begin{tcolorbox}[colback=gray!5!white, colframe=gray!75!black, title=Example 3--WNUT-17, sharp corners=southwest, breakable]
\label{appendixB-WNUT-17-1}
\small
\textbf{Task Description}

You are an advanced Named-Entity Recognition (NER) system. Your task is to analyze the given sentence or passage, identify, extract, and classify specific named entities according to the following predefined entity types:
\\

\textbf{['corporation', 'creative-work', 'group', 'location', 'person', 'product']
}
\\

For each sentence:
\begin{itemize}
    \item \textbf{Label} each word in the text with the appropriate entity type if it matches the specified categories.
    \item Extract \textbf{multiple entities} of the same class if they exist.
\end{itemize}

The output should be in \textbf{valid JSON format}, with each word and its corresponding label as shown below.

Follow the structure strictly and do not add any other explanation. In entities, label the word exactly as in the text. All the text is case-sensitive.

\subsection*{Examples}

\begin{lstlisting}[breaklines]
    
['@justinbieber i just wanna say you make me smile everyday :) thanks for smiling because when u smile i smile ! :)', []]

["@joeymcintyre I heart you . Even if I haven't seen u in months ... SEND A PIC !", []]

['@lovable_sin OMG OMG OMG ! Thank you for " tumblring " it to me , I so wasn\'t expecting them today . OMG !', []]

['RT @aplusk : This made me laugh today http://bit.ly/bjOhom &lt; --- courtesy of splurb . What made you laugh ?', []]

['RT @Sn00ki : Haha yes !!! I love that you knew that :) RT @trishamelissa @Sn00ki Is phenomenal the word of the day ?', []]

\end{lstlisting}

\subsection*{Input}

@jimmyfallon is following me ! OMG ! My life is now complete ! I heart you JF and have for years ! Thank you for making me laugh everyday !
\\

\hrule

\subsection*{Response}

[{Entity: @jimmyfallon, Label: person}, {Entity: JF, Label: person}] 
 \end{tcolorbox}


\begin{tcolorbox}[colback=gray!5!white, colframe=gray!75!black, title=Example 4--WNUT-17, sharp corners=southwest, breakable]
\label{appendixB-WNUT-17-2}
\small
\textbf{Task Description}

You are an advanced Named-Entity Recognition (NER) system. Your task is to analyze the given sentence or passage, identify, extract, and classify specific named entities according to the following predefined entity types:
\\

\textbf{['corporation', 'creative-work', 'group', 'location', 'person', 'product']
}
\\

For each sentence:
\begin{itemize}
    \item \textbf{Label} each word in the text with the appropriate entity type if it matches the specified categories.
    \item Extract \textbf{multiple entities} of the same class if they exist.
\end{itemize}

The output should be in \textbf{valid JSON format}, with each word and its corresponding label as shown below.

Follow the structure strictly and do not add any other explanation. In entities, label the word exactly as in the text. All the text is case-sensitive.

\subsection*{Examples}

\begin{lstlisting}[breaklines]
    
['We are one step closer to our new kitchens . We chose a maker and had official measurements taken today !', []]

['We were all enjoying a glass of wine in the office when a fudge delivery showed up . I love my job . And I love Fridays .', []]

['800 miles to see clients , 3 ACC candidate/commissioner meetings , big press release , making it to Friday .. PRICELESS !', []]

["I hope the weeks keep flying . It 's actually fantastic the way none of the days dragged this week .... like NONE . :D", []]

['Feeling really good after great week in our SF and LA offices . Glad to kick back on AMerican flight back to NYC', [{'Entity': 'SF', 'Label': 'location'}, {'Entity': 'LA', 'Label': 'location'}, {'Entity': 'AMerican', 'Label': 'corporation'}, {'Entity': 'NYC', 'Label': 'location'}]]

\end{lstlisting}

\subsection*{Input}

Great week in the Optimise office, another new client on board and we are close to signing a new team member
\\

\hrule

\subsection*{Response}

[{Entity: Optimise, Label: corporation}]
\end{tcolorbox}

\begin{tcolorbox}[colback=gray!5!white, colframe=gray!75!black, title=Example 5--SKILLSPAN, sharp corners=southwest, breakable]
\label{appendixB-SKILLSPAN-1}

\textbf{Task Description}

You are an advanced Named-Entity Recognition (NER) system. Your task is to analyze the given sentence or passage, identify, extract, and classify specific named entities according to the following predefined entity types:
\\

\textbf{['Skill']}
\\

For each sentence:
\begin{itemize}
    \item \textbf{Label} each word in the text with the appropriate entity type if it matches the specified categories.
    \item Extract \textbf{multiple entities} of the same class if they exist.
\end{itemize}

The output should be in \textbf{valid JSON format}, with each word and its corresponding label as shown below.

Follow the structure strictly and do not add any other explanation. In entities, label the word exactly as in the text. All the text is case-sensitive.

\subsection*{Examples}

\begin{lstlisting}[breaklines]
    
['Hands on experience with automated testing using Java .', []]

['Experience with automation systems framework design/use and deployment .', []]

['Good understanding of Agile methodologies and Continuous Delivery .', []]

['Demonstrate clear understanding of automation and orchestration principles .', []]

['Good exposure to UI Frameworks like Angular Proficiency in SQL and Database development .', []]

["Ability to understand and use efficient Defect management regular view of test coverage to identify gaps and provide improvements Personal Specification 5+ years of relevant IT/quality assurance work experience Bachelor's degree in Computer Science or related field of study or equivalent relevant experience; demonstrated experience within the quality assurance / testing arena; demonstrated skills in quality assurance methods/processes and practices .", [{'Entity': 'understand and use efficient Defect management', 'Label': 'Skill'}, {'Entity': 'identify gaps', 'Label': 'Skill'}]]

\end{lstlisting}

\subsection*{Input}

Very good understanding of test automation frameworks.
\\
\hrule
\subsection*{Response}

[{Entity: test automation frameworks, Label: Skill}]

\end{tcolorbox}


\begin{tcolorbox}[colback=gray!5!white, colframe=gray!75!black, title=Example 6--SKILLSPAN, sharp corners=southwest, breakable]
\label{appendixB-SKILLSPAN-2}
\small
\textbf{Task Description}

You are an advanced Named-Entity Recognition (NER) system. Your task is to analyze the given sentence or passage, identify, extract, and classify specific named entities according to the following predefined entity types:
\\

\textbf{['Skill']}
\\

For each sentence:
\begin{itemize}
    \item \textbf{Label} each word in the text with the appropriate entity type if it matches the specified categories.
    \item Extract \textbf{multiple entities} of the same class if they exist.
\end{itemize}

The output should be in \textbf{valid JSON format}, with each word and its corresponding label as shown below.

Follow the structure strictly and do not add any other explanation. In entities, label the word exactly as in the text. All the text is case-sensitive.

\subsection*{Examples}

\begin{lstlisting}[breaklines]
    
['Strong communication skills including the ability to express complex technical concepts to different audiences in writing and conference calls .', [{'Entity': 'communication skills', 'Label': 'Skill'}, {'Entity': 'express complex technical concepts to different audiences', 'Label': 'Skill'}]]

['Excellent organizational verbal and written communication skills .', [{'Entity': 'organizational verbal and written communication skills', 'Label': 'Skill'}]]

['Excellent organizational verbal and written communication skills .', [{'Entity': 'organizational verbal and written communication skills', 'Label': 'Skill'}]]

['The ability to work within a team and in collaboration with others is critical to this position and excellent communication skills verbal and written are essential .', [{'Entity': 'work within a team and in collaboration with others', 'Label': 'Skill'}, {'Entity': 'communication skills', 'Label': 'Skill'}]]

['This role requires a wide variety of strengths and capabilities including Ability to work collaboratively in teams and develop meaningful relationships to achieve common goals Strong organizational skills Ability to multi-task and deliver to a tight deadline Excellent written and verbal communication skills Experience developing UI components in Angular Good experience in using design patterns UML OO concepts .', [{'Entity': 'work collaboratively in teams', 'Label': 'Skill'}, {'Entity': 'develop meaningful relationships', 'Label': 'Skill'}, {'Entity': 'achieve common goals', 'Label': 'Skill'}, {'Entity': 'organizational skills', 'Label': 'Skill'}, {'Entity': 'multi-task', 'Label': 'Skill'}, {'Entity': 'deliver to a tight deadline', 'Label': 'Skill'}, {'Entity': 'communication skills', 'Label': 'Skill'}, {'Entity': 'developing UI components', 'Label': 'Skill'}, {'Entity': 'using design patterns', 'Label': 'Skill'}]]

\end{lstlisting}

\subsection*{Input}

Must have excellent verbal and written skills being able to communicate effectively on both a technical and business level Ability to work under pressure to resolve issues affecting the production services .
\\
\hrule

\subsection*{Response}

[{Entity: verbal and written skills, Label: Skill}, 
{Entity: communicate effectively on both a technical and business level, Label: Skill}, 
{Entity: work under pressure, Label: Skill}, 
{Entity: resolve issues affecting the production services, Label: Skill}]
\end{tcolorbox}


\begin{tcolorbox}[colback=gray!5!white, colframe=gray!75!black, title=Example 7--GUM, sharp corners=southwest, breakable]
\label{appendixB-GUM-1}
\small
\textbf{Task Description}

You are an advanced Named-Entity Recognition (NER) system. Your task is to analyze the given sentence or passage, identify, extract, and classify specific named entities according to the following predefined entity types:
\\

\textbf{['abstract', 'animal', 'event', 'object', 'organization', 'person', 'place', 'plant', 'quantity', 'substance', 'time']}
\\

For each sentence:
\begin{itemize}
    \item \textbf{Label} each word in the text with the appropriate entity type if it matches the specified categories.
    \item Extract \textbf{multiple entities} of the same class if they exist.
\end{itemize}

The output should be in \textbf{valid JSON format}, with each word and its corresponding label as shown below.

Follow the structure strictly and do not add any other explanation. In entities, label the word exactly as in the text. All the text is case-sensitive.

\subsection*{Examples}

\begin{lstlisting}[breaklines]
    
[ 'The 131-page document was found on Castlefrank Road in Kanata , Ontario in a rain-stained , tire-marked brown envelope by a passerby', 'Entities': [{'Entity': 'The 131-page document was found on Castlefrank Road in Kanata , Ontario in a rain-stained , tire-marked brown envelope by a passerby', 'Label': 'event'}]]

['Also the language is important in writing and in literature', 'Entities': [{'Entity': 'the language', 'Label': 'abstract'}, {'Entity': 'writing', 'Label': 'abstract'}, {'Entity': 'literature', 'Label': 'abstract'}]]

['Ingredients Basil comes in many different varieties , each of which have a unique flavor and smell', 'Entities': [{'Entity': 'Ingredients', 'Label': 'object'}, {'Entity': 'Basil', 'Label': 'plant'}, {'Entity': 'many different varieties', 'Label': 'abstract'}, {'Entity': 'each of which', 'Label': 'abstract'}, {'Entity': 'a unique flavor and smell', 'Label': 'abstract'}]]

['We do not want to just traffic in the same 24 hour news cycle', 'Entities': [{'Entity': 'We do not want to just traffic in the same 24 hour news cycle', 'Label': 'abstract'}]]

['You go through quite a bit', 'Entities': [{'Entity': 'You', 'Label': 'person'}, {'Entity': 'quite a bit', 'Label': 'quantity'}]]

\end{lstlisting}

\subsection*{Input}

If you are just visiting York for the day , using a Park and Ride [ 1 ] costs a lot less than trying to park in or near the city centre , and there are five sites dotted around the Outer Ring Road
\\
\hrule

\subsection*{Response}

[{'Entity': 'York', 'Label': 'place'}, {'Entity': 'the day', 'Label': 'time'}, {'Entity': 'a Park and Ride', 'Label': 'object'}, {'Entity': 'the city centre', 'Label': 'place'}, {'Entity': 'five sites', 'Label': 'quantity'}, {'Entity': 'the Outer Ring Road', 'Label': 'place'}]

\end{tcolorbox}


\begin{tcolorbox}[colback=gray!5!white, colframe=gray!75!black, title=Example 8--GUM, sharp corners=southwest, breakable]
\label{appendixB-GUM-2}
\small
\textbf{Task Description}

You are an advanced Named-Entity Recognition (NER) system. Your task is to analyze the given sentence or passage, identify, extract, and classify specific named entities according to the following predefined entity types:
\\

\textbf{['abstract', 'animal', 'event', 'object', 'organization', 'person', 'place', 'plant', 'quantity', 'substance', 'time']}
\\

For each sentence:
\begin{itemize}
    \item \textbf{Label} each word in the text with the appropriate entity type if it matches the specified categories.
    \item Extract \textbf{multiple entities} of the same class if they exist.
\end{itemize}

The output should be in \textbf{valid JSON format}, with each word and its corresponding label as shown below.

Follow the structure strictly and do not add any other explanation. In entities, label the word exactly as in the text. All the text is case-sensitive.

\subsection*{Examples}

\begin{lstlisting}[breaklines]
    
['" NASA Administrator Charles Bolden announces where four space shuttle orbiters will be permanently displayed at the conclusion of the Space Shuttle Program during an event commemorating the 30th anniversay of the first shuttle launch on April 12 , 2011', 'Entities': [{'Entity': 'NASA Administrator Charles Bolden', 'Label': 'person'}, {'Entity': 'where four space shuttle orbiters will be permanently displayed', 'Label': 'place'}, {'Entity': 'the conclusion of the Space Shuttle Program', 'Label': 'event'}, {'Entity': 'an event', 'Label': 'event'}, {'Entity': '30th anniversay of the first shuttle launch', 'Label': 'event'}, {'Entity': 'April 12 , 2011', 'Label': 'time'}]]

['NASA celebrated the launch of the first space shuttle Tuesday at an event at the Kennedy Space Center ( KSC ) in Cape Canaveral , Florida', 'Entities': [{'Entity': 'NASA', 'Label': 'organization'}, {'Entity': 'the launch of the first space shuttle', 'Label': 'event'}, {'Entity': 'Tuesday', 'Label': 'time'}, {'Entity': 'an event', 'Label': 'event'}, {'Entity': 'Kennedy Space Center', 'Label': 'place'}, {'Entity': 'KSC', 'Label': 'place'}, {'Entity': 'Cape Canaveral , Florida', 'Label': 'place'}]]

['Looking back : Space Shuttle Columbia lifts off on STS-1 from Launch Pad 39A at the Kennedy Space Center on April 12 , 1981', 'Entities': [{'Entity': 'Space Shuttle Columbia', 'Label': 'object'}, {'Entity': 'STS-1', 'Label': 'event'}, {'Entity': 'Launch Pad 39A', 'Label': 'place'}, {'Entity': 'Kennedy Space Center', 'Label': 'place'}, {'Entity': 'April 12 , 1981', 'Label': 'time'}]]

['At the ceremony , NASA Administrator Charles Bolden announced the locations that would be given the three remaining Space Shuttle orbiters following the end of the Space Shuttle program', 'Entities': [{'Entity': 'the ceremony', 'Label': 'event'}, {'Entity': 'NASA Administrator Charles Bolden', 'Label': 'person'}, {'Entity': 'the locations', 'Label': 'place'}, {'Entity': 'the three remaining Space Shuttle orbiters', 'Label': 'object'}, {'Entity': 'the end of the Space Shuttle program', 'Label': 'event'}]]

['On April 12 , 1981 , Space Shuttle Columbia lifted off from the Kennedy Space Center on STS-1 , the first space shuttle mission', 'Entities': [{'Entity': 'April 12 , 1981', 'Label': 'time'}, {'Entity': 'Space Shuttle Columbia', 'Label': 'object'}, {'Entity': 'Kennedy Space Center', 'Label': 'place'}, {'Entity': 'STS-1', 'Label': 'event'}, {'Entity': 'the first space shuttle mission', 'Label': 'event'}]]


\end{lstlisting}

\subsection*{Input}

Tuesday , September 22 , 2015 Discovery is undergoing decommissioning and currently being prepped for display by removing toxic materials from the orbiter
\\
\hrule

\subsection*{Response}

[{'Entity': 'Tuesday', 'Label': 'time'}, {'Entity': 'September 22 , 2015', 'Label': 'time'}, {'Entity': 'Discovery', 'Label': 'object'}, {'Entity': 'decommissioning', 'Label': 'event'}, {'Entity': 'display', 'Label': 'event'}, {'Entity': 'toxic materials', 'Label': 'substance'}, {'Entity': 'the orbiter', 'Label': 'object'}]

\end{tcolorbox}

\end{document}